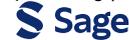

# Deep recurrent-convolutional neural network learning and physics Kalman filtering comparison in dynamic load identification

**Marios Impraimakis**

## Abstract
The dynamic structural load identification capabilities of the gated recurrent unit, long short-term memory, and convolutional neural networks are examined herein. The examination is on realistic small dataset training conditions and on a comparative view to the physics-based residual Kalman filter (RKF). The dynamic load identification suffers from the uncertainty related to obtaining poor predictions when in civil engineering applications only a low number of tests are performed or are available, or when the structural model is unidentifiable. In considering the methods, first, a simulated structure is investigated under a shaker excitation at the top floor. Second, a building in California is investigated under seismic base excitation, which results in loading for all degrees of freedom. Finally, the International Association for Structural Control-American Society of Civil Engineers (IASC-ASCE) structural health monitoring benchmark problem is examined for impact and instant loading conditions. Importantly, the methods are shown to outperform each other on different loading scenarios, while the RKF is shown to outperform the networks in physically parametrized identifiable cases.

## Keywords
Gated recurrent unit, long short-term memory artificial neural networks, deep one-dimensional convolutional networks, machine learning-intelligence, Kalman filter-based structural force identification-estimation, unknown input-load structural health monitoring

## Introduction

The four fundamental structural identification problems in civil structures[1] are (i) computing the dynamic responses with known structure parameters and loads,[2–12] (ii) solving or recovering the structural parameters based on known responses and excitations,[13–16] (iii) identifying or estimating the structure input loads using some structure parameters and responses,[17–28] and (iv) identifying or estimating the structure input loads using only the structure responses.[29–41] The first one is a forward problem, while the rest are inverse problems. While the forward problem has been researched extensively for a long time, the inverse ones, and in particular the load identification or estimation, attracted attention only recently.

The structural load, specifically, is an integral part of the system identification and monitoring process as the analytical or the numerical structural models, inevitably, are much better calibrated by input–output identification processes.[42–44] It is particularly useful for civil engineering structures since the loading is difficult to be estimated or measured due to the stochastic environment. In the input–output identification scenario, though, the required input cannot always be measured, or the measurement of the input may be more unreliable than what is demanded. For instance, there is not a reliable means of accurately measuring the traffic and wind load on large structural systems.[45–47]

To this end, several methodologies have been created to provide the structural load identification, but often they are either refer to linear systems or the methodologies are examined only on systems with no input at some degrees of freedom (DOFs). This is not the case for all civil structures; for instance, the ones which

Department of Civil, Maritime and Environmental Engineering, University of Southampton, Southampton, UK

**Corresponding author:**
Marios Impraimakis, Department of Civil, Maritime and Environmental Engineering, University of Southampton, Burgess Rd, Southampton SO16 7QF, Southampton, SO17 1BJ, UK.
Email: m.impraimakis@southampton.ac.uk

are excited at the base. Furthermore, many works examine inputs which have zero mean value, that is, white noise or some seismic excitations, excluding cases such as a hammer dynamic test scenario. As a result, these methods do not always succeed on realistic complex civil structures and a need for further investigation arises.

The importance of the dynamic structural load identification, specifically, is highlighted by the fact that a more detailed model needs to fit the parameters with even greater accuracy. This requires proper parameter sensitivity in order for the parameters to be estimated correctly. Furthermore, the load identification is also beneficial for the optimal sensor placement.[48-55] The philosophy behind those approaches is to minimize the information entropy after quantifying the uncertainty in the system parameters. This is used as a sensor configuration performance measure. The knowledge of the structural loading significantly improves the uncertainty quantification in the structural identification and, as a result, leads to a better estimation of the information gained during the model updating process.

Output-only system identification techniques, on the other hand, have also a long history of assessing the structural condition when performed during their normal operation with ambient vibration data. In this direction, the stochastic modal identification techniques are introduced from output-only data, combining high computational robustness efficiency with high estimation accuracy. To address the nonautomated identification issue in output-only procedures extensive research is performed, and it is still ongoing. Rainieri and Fabbrocino[56] presented a literature review for the most common automated output-only dynamic identification techniques. However, those methodologies are very sensitive to the noise level which often results in inaccurate estimates. This highlights the importance of identifying the loading.

To address the challenge of load identification, the deep learning architecture libraries are employed in this work. In the last few years, machine and deep learning resulted in a substantial impact on a variety of civil engineering problems,[57-59] or other problems such as visual recognition, speech recognition, and natural language processing. Among different types of deep neural networks, convolutional neural networks are studied the most.[60-74] The convolutional neural network (CNN) is a deep learning architecture inspired by the natural visual perception mechanism of the living creatures. Hubel and Wiesel[75] noticed that cells in animal visual cortex are responsible for detecting light in receptive fields. Based on this, Kunihiko Fukushima proposed the neocognitron,[76] which could be regarded as the predecessor of CNN. LeCun et al.,[77] later,

developed a multilayer artificial neural network called LeNet-5 which could classify handwritten digits.[77]

To overcome the shortcoming of deep neural networks of being difficult to be trained[78,79] due to the exploding-vanishing gradient issue when learning long-term dependencies, the long short-term memory (LSTM) architecture[80] is introduced. Importantly, the LSTM network is designed to capture long-range data dependencies on modeling sequential data such as the dynamic load, and shows a great potential in modeling structural loading time series,[81] or in other applications.[6,82,83]

In the same direction, the gated recurrent unit (GRU) neural networks[84] have shown success in several applications involving sequential or temporal data.[85] GRU success is attributed to the gating network signaling. This controls how the present input and previous memory is used to update the current activation and produce the current state. These gates have their own sets of weights which are adaptively updated in the learning phase.

Intelligent methods for dynamic load identification[86] currently focus on vehicle loads,[87-89] component and mechanical structures loading,[81,90-103] bridge cables loading,[104] and power loads.[105] They have been recently investigated in structural dynamics, but with pseudo-experimental data at the Pirelli Tower in Milan, Italy.[106]

In this work which focuses on full scale building structures with real experimental data, the structural response and loading are employed to train the neural networks, and finally predict unseen loading data. In doing so, the work contributes to the dynamic load identification research assessing the networks in the uncertain outcome related to obtaining poor predictions when in civil engineering applications only a low number of tests are performed or are available, or when the structural system is unidentifiable with physical parameter-based modeling. The networks are compared when overcoming those issues, while Kalman filter physics-based alternatives, without assuming more information such as known system parameter of the model,[41] are also employed. A realistic small dataset scenario prone to outliers is investigated for civil engineering applications in contrast to hundred or even thousands of available data which are assumed for other applications. This problem is crucial since it potentially leads to overfitting the model when it is adjusted excessively to the training data, seeing patterns that do not exist, and consequently performing poorly in predicting new data.

The work is organized as follows: the LSTM network is overviewed in section "Dynamic load identification using the LSTM neural networks," while in section "Dynamic load identification using the GRU neural networks," the improved and faster GRU is presented. The standard CNN architecture is provided in section "Dynamic load identification using 1D CNNs," as well as a discussion on the one-dimensional and the multidimensional CNN versions. The physics-based RKF is then presented in section "Dynamic load identification using physics-based residual Kalman filtering." Importantly, sections "Structural loading identification in a 6-story building," "Structural loading identification for a hotel in San Bernardino," and "Structural loading identification in the IASC-ASCE structural health monitoring benchmark problem" investigate applications on both simulated and real structures, as well as on both continuous and impact loading cases. Subsequently, section "Discussion" presents a discussion and section "Future research" future research suggestions. Finally, the conclusions are provided in section "Conclusion."

## Dynamic load identification using the LSTM neural networks

The LSTM neural networks are a type of recurrent neural networks (RNNs) which are designed to handle the vanishing gradient problem in the traditional RNNs. The vanishing gradient problem occurs when the gradients of the error function with respect to the weights in the RNN become very small. This makes it difficult for the network to learn long-term dependencies.

The standard LSTM architecture consists of several memory cells that can store information for long periods of time, as well as several gates which regulate the flow of information into and out of the cells, seen in Figure 1. The gates are controlled by sigmoid activation functions and can either allow or prevent information from passing through. The LSTM cell has three gates: the forget gate, the input gate, and the output gate. The forget gate determines which information to discard from the previous cell state, the input gate determines which new information to add to the current cell state, and the output gate determines which information to output from the current cell state. The equations governing the LSTM cell are written as

$$
\begin{aligned}
f_t &= \sigma_g\left(W_f x_t + U_f h_{t-1} + b_f\right) \\
i_t &= \sigma_g\left(W_i x_t + U_i h_{t-1} + b_i\right) \\
o_t &= \sigma_g\left(W_o x_t + U_o h_{t-1} + b_o\right) \\
\tilde{c}_t &= \sigma_c\left(W_c x_t + U_c h_{t-1} + b_c\right) \\
c_t &= f_t \odot c_{t-1} + i_t \odot \tilde{c}_t \\
h_t &= o_t \odot \sigma_h(c_t)
\end{aligned}
\tag{1}
$$

where $W_f, W_i, W_o, W_c, U_f, U_i, U_o U_c$, are the weight matrices, $b_f, b_i, b_o, b_c$ are the bias vectors, and $\sigma$ is the sigmoid function. The initial values are $c_0 = 0$ and $h_0 = 0$, the operator $\odot$ denotes the Hadamard product, and the subscript $t$ indexes the time step. Additionally, $x_t \in \mathbb{R}^d$ is the input vector to the LSTM unit, $f_t \in (0, 1)^h$ is the forget gate's activation vector, $i_t \in (0, 1)^h$ is the input/update gate's activation vector, $o_t \in (0, 1)^h$ is the output gate's activation vector, $h_t \in (-1, 1)^h$ is the hidden state vector also known as output vector of the LSTM unit, $\tilde{c}_t \in (-1, 1)^h$ is the cell input activation vector, $c_t \in \mathbb{R}^h$ is the cell state vector, and the superscript $h$ refers to the number of hidden units. Finally, $\sigma_g$ is the sigmoid function, while $\sigma_c$ and $\sigma_h$ are the hyperbolic tangent functions.

To train a LSTM neural network, the structural load and response set of input–output pairs are provided, also known as training data. During training, the network adjusts its weights and biases to minimize a loss function, which measures the difference between the predicted and actual output values. The loss function used to train the LSTM depends on the specific task. In the context of predicting structural loading based on the structure response, a common choice is the mean squared error between the predicted and actual loading values, mathematically written as

$$
L = \frac{1}{N} \sum_{i=1}^{N} \left(y_i - y_{i,\text{true}}\right)^2
\tag{2}
$$

where $N$ is the number of samples, $y_i$ is the predicted loading value at position $i$, and $y_{i,\text{true}}$ is the true loading value at position $i$. The LSTM network is finally trained using backpropagation through time, which involves computing the gradient of the loss function with respect to the weights and biases at each time step. These are adjusted using an optimization algorithm such as the stochastic gradient descent.

## Dynamic load identification using the GRU neural networks

GRU neural networks, on the other hand, are another type of RNN which, similar to the LSTM case, are designed to handle the vanishing gradient problem in traditional RNNs. GRUs are similar to LSTMs in that they also use gating mechanisms to regulate the flow of information. They are simpler and more computationally efficient, though.

The GRU architecture also consists of memory cells that can store information for long periods of time, as well as several gates that regulate the flow of information into and out of the cells (Figure 1). The gates are controlled by sigmoid activation functions and can either allow or prevent information from passing through. The GRU cell has two gates: the reset gate

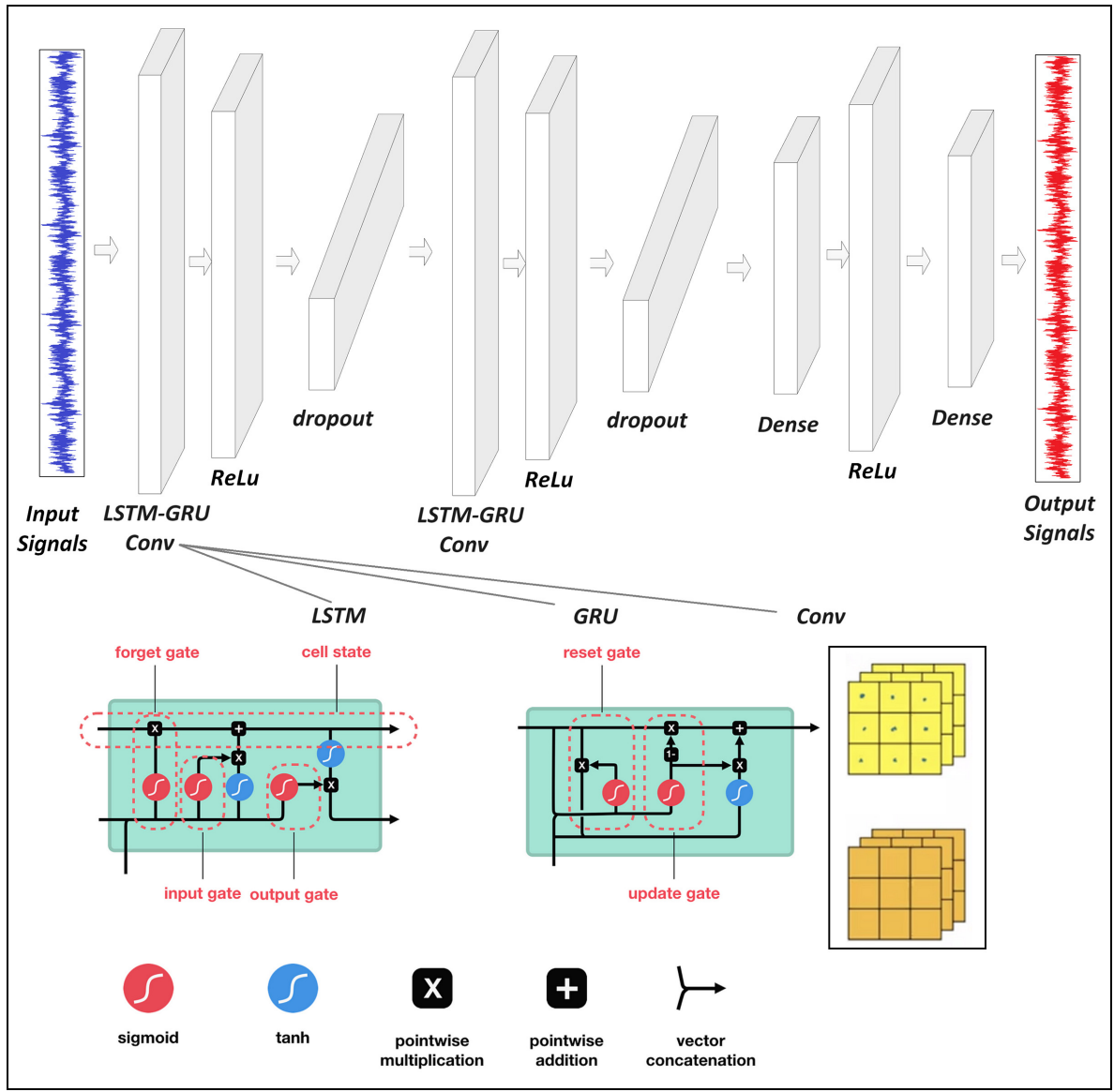

**Figure 1.** Examined network architecture for all applications as described in section "Structural loading identification in a 6-story building." The LSTM–GRU–Conv layer is replaced each time by each one of the three considered layers. The dropout layer is removed in the 1D-CNN case.

LSTM: long short-term memory artificial neural network; GRU: gated recurrent unit; 1D-CNN: one-dimensional convolutional neural network.

and the update gate. The reset gate determines how much of the previous state to forget, while the update gate determines how much of the new state to add to the current state. The equations governing the GRU cell are written as

$$z_t = \sigma(W_z[h_{t-1}, x_t] + b_z)$$
$$r_t = \sigma(W_r[h_{t-1}, x_t] + b_r)$$
$$h'_t = \tanh(W_h[r_t \odot h_{t-1}, x_t] + b_h) \quad (3)$$
$$h_t = (1 - z_t) \odot h_{t-1} + z_t \odot h'_t$$

where $x_t$ is the input vector, $h_t$ is the output vector, $h'_t$ is the candidate activation vector, $z_t$ is the update gate vector, $r_t$ is the reset gate vector, and $W$ and $b$ are the parameter matrices and vector, respectively. Finally, $\sigma$ is the original logistic function.

To train a GRU neural network, the structural load and response set of input–output pairs are also provided. It adjusts its weights and biases to minimize a loss function which measures the difference between the predicted and actual output values. The loss function used to train the GRU also depends on the specific task. In the context of predicting structural loading based on structural response, a common choice is the mean squared error between the predicted and actual loading signals. The GRU network is trained using backpropagation through time, similar to LSTMs, which involves computing the gradient of the loss function with respect to the weights and biases at each time step.

## Dynamic load identification using 1D CNNs

Finally, the one-dimensional convolutional neural networks (1D CNNs) have been proven to be highly effective in a variety of signal processing tasks. The fundamental building block of a 1D CNN is the convolutional layer (Figure 1). A convolutional layer applies a set of filters to the input signal, producing a set of feature maps. The filters have a fixed size and slide over the input signal, computing a dot product at each location. The resulting feature maps capture different aspects of the input signal, such as local trends and patterns.

The applied 1D CNN compares to the multidimensional counterparts as follows: A one-dimensional configuration fuses the feature extraction and the learning phases of the dynamic states. One-dimensional arrays are used instead of two-dimensional matrices for both the kernels and the feature maps. Additionally, the network architecture has the hidden neurons of the convolution layers which perform both the convolution and the subsampling operations. Accordingly, the convolution and the lateral rotation are replaced by their one-dimensional counterparts, namely the convolution and the reverse operations. Finally, the parameters for the kernel size and the subsampling are scalars. Importantly, this simplified structure of the convolution neural network requires only one-dimensional convolutions and therefore, a mobile and low-cost hardware implementation for near real-time applications. The convolution operation is represented mathematically as

$$h_i = f\left(\sum_{j=0}^{m-1} w_j x_{i+j} + b\right) \tag{4}$$

where $h_i$ is the output at position $i$, $w_j$ is the weight of the $j$th filter, $x_{i+j}$ is the input signal at position $i+j$, $b$ is the bias term, and $f$ is the activation function.

In practice, a 1D CNN may have multiple convolutional layers with different filter sizes and numbers of filters. Each layer can apply a different set of filters to the input signal, allowing the network to capture different aspects of the signal at different scales. No additional layers are assumed in this work for the 1D CNN (such as pooling layers) to compare fairly all networks.

To train the 1D CNN, the structural load and response set of input–output pairs are also provided. During training, the network adjusts its weights and biases to minimize the loss function, which measures the difference between the predicted and actual loading values. This is done using an optimization algorithm such as the stochastic gradient descent. This updates the weights and biases based on the gradient of the loss function.

For all three networks discussed in this work, in addition to the training data, it is important to have a separate set of validation data to monitor the training performance of the network to avoid overfitting. The validation data are used to evaluate the network's performance on unseen data, and the training process can be stopped early if the performance on the validation data starts to deteriorate.

## Dynamic load identification using physics-based residual Kalman filtering

For the mathematical implementation of the unknown input residual-based Kalman filter[41] consider the process equation in the continuous-time and the state-space format:

$$\dot{\mathbf{z}} = \mathbf{A}\mathbf{z} + \mathbf{B}\mathbf{u} \tag{5}$$

where $\mathbf{A}(\boldsymbol{\theta})$ is the system matrix depended on the unknown parameter vector $\boldsymbol{\theta}$, $\mathbf{B}$ is the distribution matrix of the input $\mathbf{u}$, and $\mathbf{z}$ is the dynamic state vector.

The discrete-time transformation of the system and the input matrices is provided by the zero-order hold assumption for the input in between the time instants $k\Delta t$, as:

$$\mathbf{A_d} = e^{\mathbf{A}\Delta t} \approx \mathbf{I_{2n \times 2n}} + \Delta t \mathbf{A} + \frac{\Delta t}{2}\mathbf{A}^2 \tag{6}$$

and

$$\mathbf{B_d} = \int_0^{\Delta t} e^{\mathbf{A}\tau}\mathbf{B}\mathrm{d}\tau = \mathbf{A}^{-1}[\mathbf{A_d} - \mathbf{I_{2n \times 2n}}]\mathbf{B} \approx \Delta t \mathbf{B} \tag{7}$$

The state-space model of Equation (5) in the discrete-time, including the noise term $\mathbf{w_k}$, is written as

$$\mathbf{z_{k+1}} = \mathbf{A_d}\mathbf{z_k} + \mathbf{B_d}\mathbf{u_k^e} + \mathbf{w_k} \tag{8}$$

where $\mathbf{u_k^e}$ and $\mathbf{A_d}(\boldsymbol{\theta_k})$ are the estimated input and system matrix of the prior step which is considered as known quantities at the $k+1$ step.

The equation which relates the measurements $\mathbf{y}$ to the estimated dynamic states is written as

$$\mathbf{y}_{k+1} = \mathbf{H}\mathbf{z}_{k+1} + \mathbf{w}_{k+1}^{y} \tag{9}$$

where $\mathbf{H}$ is the observation matrix mapping the measurements to the dynamic states. Here, it is chosen to not depend on the unknown parameters and input. To this end and for limited information applications, $\mathbf{y}$ consists of displacement and velocity pseudo-measurements; the integrated of the actual acceleration measurements. Additionally, the accelerations which are not measured are assumed to be equal to the estimated accelerations of the previous step. Specifically, the matrix $\mathbf{H}$ is introduced as the observation matrix mapping measurements to dynamical states. This matrix only accommodates displacements and velocities observed from all DOFs with real or pseudo-measurements.

It may seem here that the acceleration responses are not covered by the observation matrix. However, this is chosen intentionally since it addressees two problems. First, the unknown input and parameters have not yet been estimated for the step $k+1$. Second, the prior step parameters and input possibly affect negatively the observation equation when they are inaccurate.

More importantly, the presented observation model reflects the model for the pseudo-measurements rather than the actual measured quantities. In that case, the actual observation model, which relates the observed quantities to the state vector, is not defined. To clarify how different measurement scenarios are accommodated within this approach and at which step they weigh in, the reader is referred to Impraimakis and Smyth.[41]

The predicted covariance matrix $\mathbf{P}_{k+1}$ of the dynamic states is then written as

$$\mathbf{P}_{k+1} = \mathbf{A}_d \mathbf{P}_k \mathbf{A}_d^{T} + \mathbf{Q}_{d(k)} \tag{10}$$

where the discretized process and observation covariance matrices are

$$\mathbf{Q}_{k-1} \approx \frac{\mathbf{Q}((k-1)\Delta t)}{\Delta t}, \ \mathbf{R}_k = \frac{\mathbf{R}(k\Delta t)}{\Delta t} \tag{11}$$

It is assumed, though, that the matrices are constant during the whole process, where being constant does not harm the estimation success. An investigation of their exact value, which importantly highly affects the success of the estimation,[107] is shown in Refs. 41 and 108.

Having provided the posterior prediction model for the dynamic states and their covariances, the update process starts according to the Kalman filter. The updated dynamic state estimate is specifically derived by a correction of the predicted dynamic states using the measurement pre-fit residual. This is multiplied and controlled by the optimal Kalman gain $\mathbf{J}$, given as

$$\mathbf{J}_{k+1} = \mathbf{P}_{k+1}\mathbf{H}^{T}\mathbf{N}_{k+1}^{-1} \tag{12}$$

where the pre-fit residual covariance $\mathbf{N}$ is

$$\mathbf{N}_{k+1} = \mathbf{H}\mathbf{P}_{k+1}\mathbf{H}^{T} + \mathbf{R}_d \tag{13}$$

The final estimation of the posterior dynamic states is then given by

$$\mathbf{z}_{k+1} = \mathbf{z}_{k+1} + \mathbf{J}_{k+1}(\mathbf{y}_{k+1} - \mathbf{H}\mathbf{z}_{k+1}) \tag{14}$$

while the final estimation of the covariance of the dynamic states is given by

$$\mathbf{P}_{k+1} = (\mathbf{I}_{n \times n} - \mathbf{J}_{k+1}\mathbf{H})\mathbf{P}_{k+1} \tag{15}$$

For Equations (14) and (15), the same quantity on the right and left hand side implies that they are recalculated at the same time step. The a priori estimate of the right hand side is used for the calculation of the a posteriori estimate on the left hand side.

Once the dynamic states are filtered using the pseudo-measurements and with the use of the parameters of the prior step, the input at the current step is approximated by the system model at the time instant $(k+1)\mathrm{d}t$ as

$$\mathbf{u}_{k+1}^{e} \approx \mathbf{G}(\ddot{\mathbf{x}}_{k+1}^{m}, \ \mathbf{z}_k, \ \boldsymbol{\theta}_k) \tag{16}$$

where $\mathbf{G}(\bullet)$ is the linear or nonlinear system model, which contains the prior step estimated parameters. Importantly, the predicted states are estimated using Equation (14); with the prior input and parameters only. The known input rows of $\mathbf{u}_{k+1}^{e}$ are replaced by the potential known zero or nonzero valued inputs.

For instance, the full expression of $\mathbf{G}(\bullet)$ function for a linear structural system model is written as

$$\mathbf{G}(\ddot{\mathbf{x}}_{k+1}^{m}, \ \mathbf{z}_k, \ \boldsymbol{\theta}_k) = \mathbf{M}^{m}\mathbf{a}_{k+1} + [\mathbf{K}_k \ \mathbf{C}_k]\mathbf{z}_{k+1} \tag{17}$$

where ${}^{m}a_{k+1}$ are the acceleration measurements, and $\mathbf{M}$, $\mathbf{K}_k$, $\mathbf{C}_k$ are the mass, stiffness, and damping matrices, respectively.

For the parameter estimation, a sensitivity analysis approach is implemented by the Taylor series expansion truncated after the linear term. To provide a real-time estimation specifically, the measured outputs are chosen to be accelerations instead of the modal parameters, written as

$$\boldsymbol{\epsilon}_{k+1} = {}^{m}\mathbf{a}_{k+1} - \mathbf{a}_{k+1} \approx \mathbf{r}_{k+1} + \mathbf{U}_{k+1}(\boldsymbol{\theta} - \boldsymbol{\theta}_{k+1}) \tag{18}$$

where $\boldsymbol{\epsilon}_{k+1}$, ${}^{m}\mathbf{a}_{k+1}$, and $\mathbf{a}_{k+1}$ denote the error, the acceleration measurements, and the predicted output, respectively, at the step $k + 1$. The sensitivity matrix $\mathbf{U_{k+1}}$, which does not need an initial value or prior information, is written as

$$\mathbf{U_{k+1}} = -\left[\frac{\partial^{\mathbf{m}}\mathbf{a_{k+1}}}{\partial\boldsymbol{\theta}}\right]_{\boldsymbol{\theta}=\boldsymbol{\theta}_{k+1}} \quad (19)$$

where the error $\boldsymbol{\epsilon_{k+1}}$ is assumed to be small for the parameter vector $\boldsymbol{\theta}$ in the vicinity of $\boldsymbol{\theta_{k+1}}$.

At each step, Equation (18) is solved by a Gauss–Newton gradient approach. The prior parameter estimates are corrected as

$$\boldsymbol{\theta_{k+1}} = \boldsymbol{\theta_k} + \Delta\boldsymbol{\theta_{k+1}} \cdot e^{-\mu\|\boldsymbol{\rho_{k+1}}\|_2} \quad (20)$$

where $\mu$ is a scaling parameter and $\|\boldsymbol{\rho_{k+1}}\|_2$ is the Euclidean norm of the residual of the system model estimation. In practice, $e^{-\mu\|\boldsymbol{\rho_{k+1}}\|_2}$ acts as a control factor for the convergence speed and fluctuation range. An investigation of this scaling parameter is shown in Refs. 41 and 108. A similar investigation can be done to define it for different types of model parameters within various dynamic systems.

For Equation (20), the residual of the system model estimation is

$$\boldsymbol{\rho_{k+1}} = \mathbf{u_{k+1}^e} - \mathbf{G}\left(\ddot{\mathbf{x}}_{\mathbf{k+1}}^{\mathbf{m}},\ \dot{\mathbf{x}}_{\mathbf{k+1}},\ \mathbf{x}_{\mathbf{k+1}},\ \boldsymbol{\theta}_k\right) \quad (21)$$

where $\mathbf{u_{k+1}^e}$ is the estimated input for the step $k + 1$, and the dynamic states are provided by the Kalman filter.

For the objective function, the least square approach is formulated based on an additional scaling parameter $\lambda^2$. This balances the contribution of the parameter estimates. The final optimal $\Delta\boldsymbol{\theta_{k+1}}$ correction is provided by

$$\Delta\boldsymbol{\theta_{k+1}} = \left[\mathbf{U_{k+1}^T}\mathbf{U_{k+1}} + \lambda^2\mathbf{I}\right]^{-1}\mathbf{U_{k+1}^T}\boldsymbol{\rho_{k+1}} \quad (22)$$

where $\lambda^2$ remains constant during the real-time procedure. An investigation of this scaling parameter is shown in Refs. 41 and 108. A similar investigation can be done to set both scaling parameters for different types of model parameters within various dynamic systems. Importantly, it is seen here that the transitional model assumed for the system parameters is involved in the full input-parameter-state estimation. Taking partial derivatives is then required. Also, the scaling factor is tied to the difference between the estimated and the predicted input forces. The nature, the order of magnitude, and the governing equations for the input and model parameters are different, but this approach shows to be beneficial in yielding stable estimates for the model parameters.

Regarding the derivation process of Equation (22), it is provided as the optimal solution of the objective function minimization. Here, the scaling parameter $\lambda^2$, which balances the contribution of the parameter estimation and the importance of the error $\boldsymbol{\epsilon}$, is written as

$$\mathbf{F}(\boldsymbol{\theta_{k+1}}) = \boldsymbol{\epsilon}_{k+1}^T\mathbf{W_\epsilon}\boldsymbol{\epsilon_{k+1}} + \lambda^2\Delta\boldsymbol{\theta}_{k+1}^T\mathbf{W_\theta}\Delta\boldsymbol{\theta_{k+1}} \quad (23)$$

where a penalization exists for the differences between the estimated parameters and the output error. Further derivation details are provided in Impraimakis and Smyth.[41]

## Structural loading identification in a 6-story building

For the numerical application of the GRU network, LSTM network, convolutional network, and residual-based Kalman filter for structural load identification with small datasets, consider the 6-story shear-type structure of Figure 2. The structure is described by the following equation:

$$\mathbf{M}\begin{Bmatrix}\ddot{y}_1(t)\\\ddot{y}_2(t)\\\ddot{y}_3(t)\\\ddot{y}_4(t)\\\ddot{y}_5(t)\\\ddot{y}_6(t)\end{Bmatrix} + \mathbf{C}\begin{Bmatrix}\dot{y}_1(t)\\\dot{y}_2(t)\\\dot{y}_3(t)\\\dot{y}_4(t)\\\dot{y}_5(t)\\\dot{y}_6(t)\end{Bmatrix} + \mathbf{K}\begin{Bmatrix}y_1(t)\\y_2(t)\\y_3(t)\\y_4(t)\\y_5(t)\\y_6(t)\end{Bmatrix} = \begin{Bmatrix}0\\0\\0\\0\\0\\F_6(t)\end{Bmatrix} \quad (24)$$

for a shaker-type load input $F_6(t)$ at the top floor, namely at DOF 6, where the structure matrices to generate the simulated response measurements are

$$\mathbf{M} = \begin{bmatrix}m_1 & 0 & 0 & 0 & 0 & 0\\0 & m_2 & 0 & 0 & 0 & 0\\0 & 0 & m_3 & 0 & 0 & 0\\0 & 0 & 0 & m_4 & 0 & 0\\0 & 0 & 0 & 0 & m_5 & 0\\0 & 0 & 0 & 0 & 0 & m_6\end{bmatrix} = \begin{bmatrix}100 & 0 & 0 & 0 & 0 & 0\\0 & 100 & 0 & 0 & 0 & 0\\0 & 0 & 100 & 0 & 0 & 0\\0 & 0 & 0 & 100 & 0 & 0\\0 & 0 & 0 & 0 & 100 & 0\\0 & 0 & 0 & 0 & 0 & 100\end{bmatrix},$$

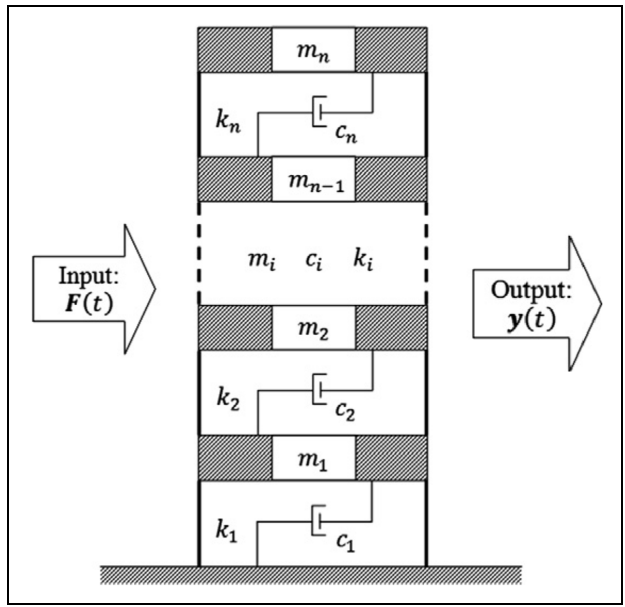

**Figure 2.** 6-story shear-type building structure of section "Structural loading identification in a 6-story building."

$$
\mathbf{C} = \begin{bmatrix}
c_1 + c_2 & -c_2 & 0 & 0 & 0 & 0 \\
-c_2 & c_2 + c_3 & -c_3 & 0 & 0 & 0 \\
0 & -c_3 & c_3 + c_4 & -c_4 & 0 & 0 \\
0 & 0 & -c_4 & c_4 + c_5 & -c_5 & 0 \\
0 & 0 & 0 & -c_5 & c_5 + c_6 & -c_6 \\
0 & 0 & 0 & 0 & -c_5 & c_6
\end{bmatrix}
$$

$$
= \begin{bmatrix}
25 + 25 & -25 & 0 & 0 & 0 & 0 \\
-25 & 25 + 50 & -50 & 0 & 0 & 0 \\
0 & -50 & 50 + 50 & -50 & 0 & 0 \\
0 & 0 & -50 & 50 + 75 & -75 & 0 \\
0 & 0 & 0 & -75 & 75 + 75 & -75 \\
0 & 0 & 0 & 0 & -75 & 75
\end{bmatrix},
$$

$$
\mathbf{K} = \begin{bmatrix}
k_1 + k_2 & -k_2 & 0 & 0 & 0 & 0 \\
-k_2 & k_2 + k_3 & -k_3 & 0 & 0 & 0 \\
0 & -k_3 & k_3 + k_4 & -k_4 & 0 & 0 \\
0 & 0 & -k_4 & k_4 + k_5 & -k_5 & 0 \\
0 & 0 & 0 & -k_5 & k_5 + k_6 & -k_6 \\
0 & 0 & 0 & 0 & -k_5 & k_6
\end{bmatrix}
$$

$$
= \begin{bmatrix}
900 + 900 & -900 & 0 & 0 & 0 & 0 \\
-900 & 900 + 1100 & -1100 & 0 & 0 & 0 \\
0 & -1100 & 1100 + 1100 & -1100 & 0 & 0 \\
0 & 0 & -1100 & 1100 + 1300 & -1300 & 0 \\
0 & 0 & 0 & -1300 & 1300 + 1300 & -1300 \\
0 & 0 & 0 & 0 & -1300 & 1300
\end{bmatrix}
$$

with initial conditions $\mathbf{y}(0) = [0 \quad 0 \quad 0 \quad 0 \quad 0 \quad 0]^T$ and $\dot{\mathbf{y}}(0) = [0 \quad 0 \quad 0 \quad 0 \quad 0 \quad 0]^T$. The input load at floor six is a harmonic loading decaying exponentially with various amplitude levels, angular frequencies, and unknown time instant of application.

In order to create synthetic measurements, the Runge Kutta fourth order method of integration is utilized to compute the system response for 200 s. The sampling frequency for the dynamic state measurements is considered to be 100 Hz. Therefore, the time discretization $\Delta t$ used in the Runge–Kutta numerical solution is 0.01 s. Finally, to consider measurement noise, each response signal is contaminated by a Gaussian white noise sequence with a 5% root-mean-square noise-to-signal ratio. Different levels of noise are investigated in section "Discussion."

A total of 21 available datasets from the simulations are formatted and divided into three subsets, including 11 datasets for training, four datasets for validation, and six datasets for prediction of the structural loading. For the Kalman filter, the identification is performed in real time, without any training. Importantly, for the shear-type building study, despite being numerical, the datasets for training, validation and test are so small, for instance, only 11 datasets for training, to match and directly compare to section "Structural loading identification for a hotel in San Bernardino" case, where also 11 datasets for training are used.

The neural network architectures are defined as follows in Figure 1: An input layer with the 11 signals for each one of the three network types. A GRU, or a LSTM or a convolutional layer with 30 units. Therefore, the dimension of the output vector is 30, while the batch-size equals to 2. A rectifier layer, termed also as ReLu is also set, as well as a dropout layer of 0.3 for the first two networks. An additional GRU, or a LSTM or a convolutional layer with 30 units is set with an additional activation layer and dropout layer for the first two cases. Finally, 100 neural density is defined for all cases, along with activation and dense layers. The learning rate is defined as 0.0001. The Adaptive Momentum Estimation (Adam) algorithm is used for the network optimization[109] and the number of epochs is 10,000. It is generally known that the performance of deep neural network is overly dependent on the setting of hyperparameters. The author set the network parameters according to Kingma and Ba[109] without any special adjustments that would potentially favor the dynamic load identification problem. Importantly, this architecture and the number of hidden units were selected as they have been proven efficient in a number of structural engineering applications.[6,110,111] Last but not least, investigation on the dropout layer hyperparameter, or the number of layers is shown in section "Discussion."

For the RKF, the process covariance $\mathbf{Q_d}$ and the measurement covariance $\mathbf{R_d}$ matrices are chosen to be constant during the identification process and equal to $10^0 \cdot \mathbf{I_{6\times6}}$ and $10^{-10} \cdot \mathbf{I_{6\times6}}$, respectively. The parameter $\lambda^2$ is chosen to be $5\times10^{-2}$, while the parameter $\mu$ is chosen to be $5\times10^{-3}$.

All three cases are examined on Figures 3 to 5. They show the true and identified structural load (first column) for the six unknown predicted datasets where the network never trained or validated. The load identification error is also seen at floor six (second column), as well as the comparison to the Kalman filter (third column). In all cases, acceleration measurement are only selected from story 3, 5, and 6. For a different combination or number of measurements, different convergence timing is observed.

Figure 3 refers to the case where the LSTM neural network is used. The results are satisfactory for all six cases. The exemption of the first loading instances is related to the slightly wrong estimation of the load phase or amplitude.

Figure 4 refers to the case where the GRU neural network is used. The results are also satisfactory for all six cases. However, a convergence time improvement is seen compared to the LSTM neural network case, as discussed in Table 2 of section "Discussion." Once more, but at a lower level, the first loading instances are not satisfactory for the same reasons as in Figuer 3.

Figure 5 refers to the case where the CNN is used. The results for all six cases are not as satisfactory as with the previous networks. However, a clear and significant convergence time reduction is observed. Importantly, for all cases, the Kalman filter approach provided a better accuracy.

So far, only time-historical error is provided to indicate the performance of presented approaches. To use a comprehensive evaluation metric, the accumulated error at each time instant is employed as

$$E(t) = \sum_{t_k=0}^{t} \left| \frac{u_{\text{pred}}(t_k) - u_{\text{true}}(t_k)}{u_{\text{true}}(t_k)} \right| \qquad (25)$$

where $t_k$ is the time instant at step $k$, and $u_{\text{pred}(t_k)}$ and $u_{\text{true}}(t_k)$ are the predicted and true load at $t_k$.

Figure 6 refers to the case where the error metric $E(t)$ of Equation (25) is used. The results show that the LSTM network performs better but with higher computation cost, which is shown in Table 2 of section "Discussion." The Kalman filter has the lower error, summarized also in Table 1.

## Structural loading identification for a hotel in San Bernardino

The methodologies are examined also in field sensing data. An examination is conducted on a 6-story hotel building in San Bernardino, California, sourced from the Center for Engineering Strong Motion Data.[112] The structure, a mid-rise concrete building designed in 1970, is equipped with nine accelerometers on the 1st floor, 3rd floor, and the roof level, as depicted in Figure 7. These sensors have captured seismic events from 1987 to 2018. The historical data serve as training inputs for the proposed neural network deep learning models, predicting the structural loading induced by the ground motions. In this scenario, methodologies such as the Kalman filter are vulnerable to identifiability issues, and they cannot be used efficiently to recover the input without assuming any known model parameter.[113] Assuming known parameters leads to unfair comparison with the network as more information is provided. For that case, the load estimation unsurprisingly is better as already demonstrated in Eftekhar Azam et al.[114]

In this examination, the field data, characterized by varying sampling rates and high-frequency noise, undergo initial preprocessing involving resampling at 100 Hz and filtering. A total of 21 datasets are organized into three subsets: 11 for training, 4 for validation, and 6 for prediction. The seismic loading at the building base is considered over a duration of 70 s. Importantly, the neural network architectures are structured in a manner consistent to the approach detailed in section "Structural loading identification in a 6-story building."

All three cases are examined on Figures 8 to 10. They show the true and identified structural load (first column) for the six unknown predicted datasets where the network never trained or validated. The load identification error at the building base is shown on the second figure column. In all cases, acceleration measurement are only selected from stories 3 and 6. For a different combination or number of measurements, different convergence timing is observed.

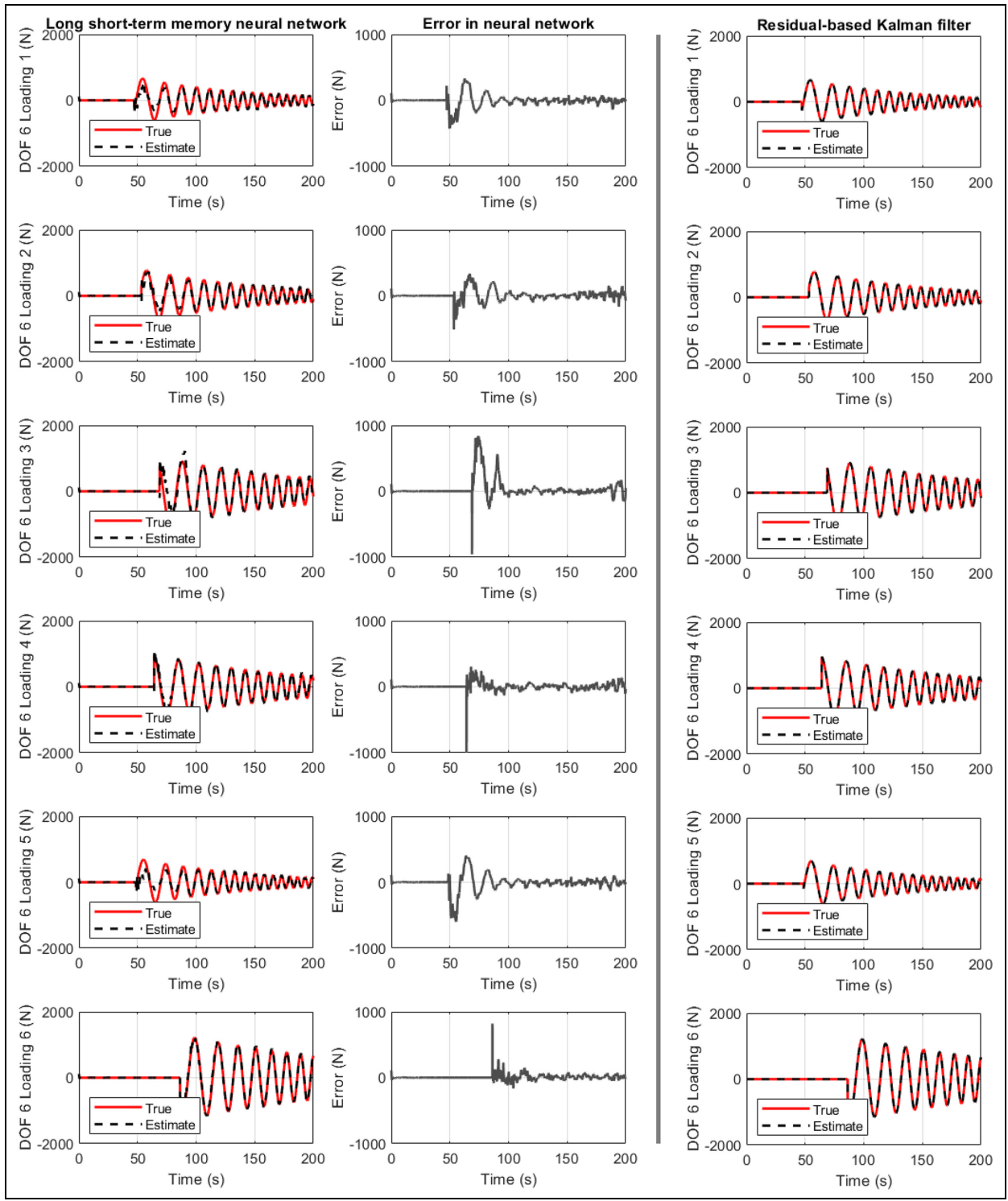

**Figure 3.** Structure of section "Structural loading identification in a 6-story building": results for the 6-story shear-type building when the LSTM neural network is used. First column: true and estimated loading at floor 6. Second column: error at loading identification. Third column: Residual-based Kalman filter performance.

LSTM: long short-term memory.

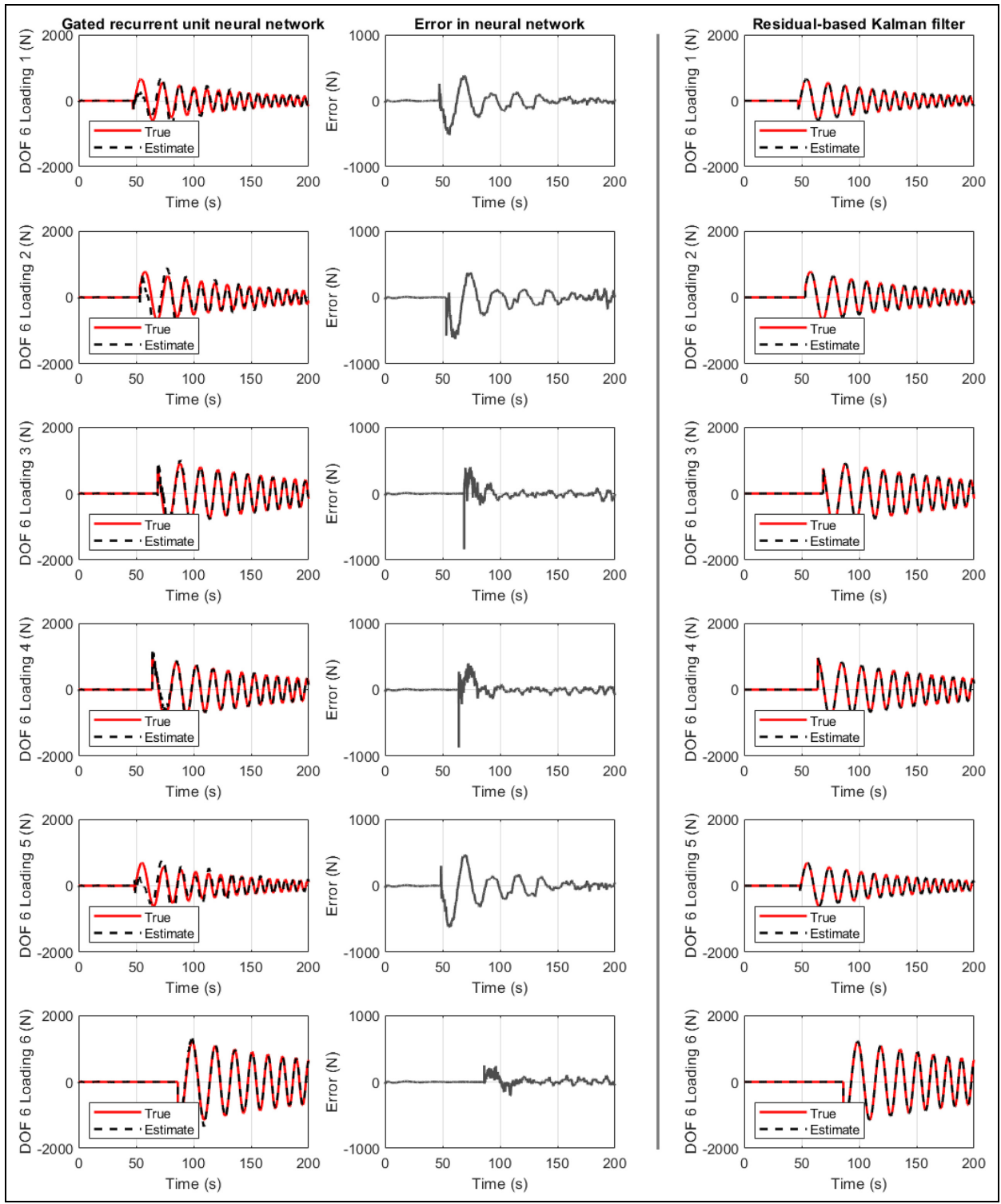

**Figure 4.** Structure of section "Structural loading identification in a 6-story building": results for the 6-story shear-type building when the gated recurrent unit neural network is used. First column: true and estimated loading at floor 6. Second column: error at loading identification. Third column: Residual-based Kalman filter performance.

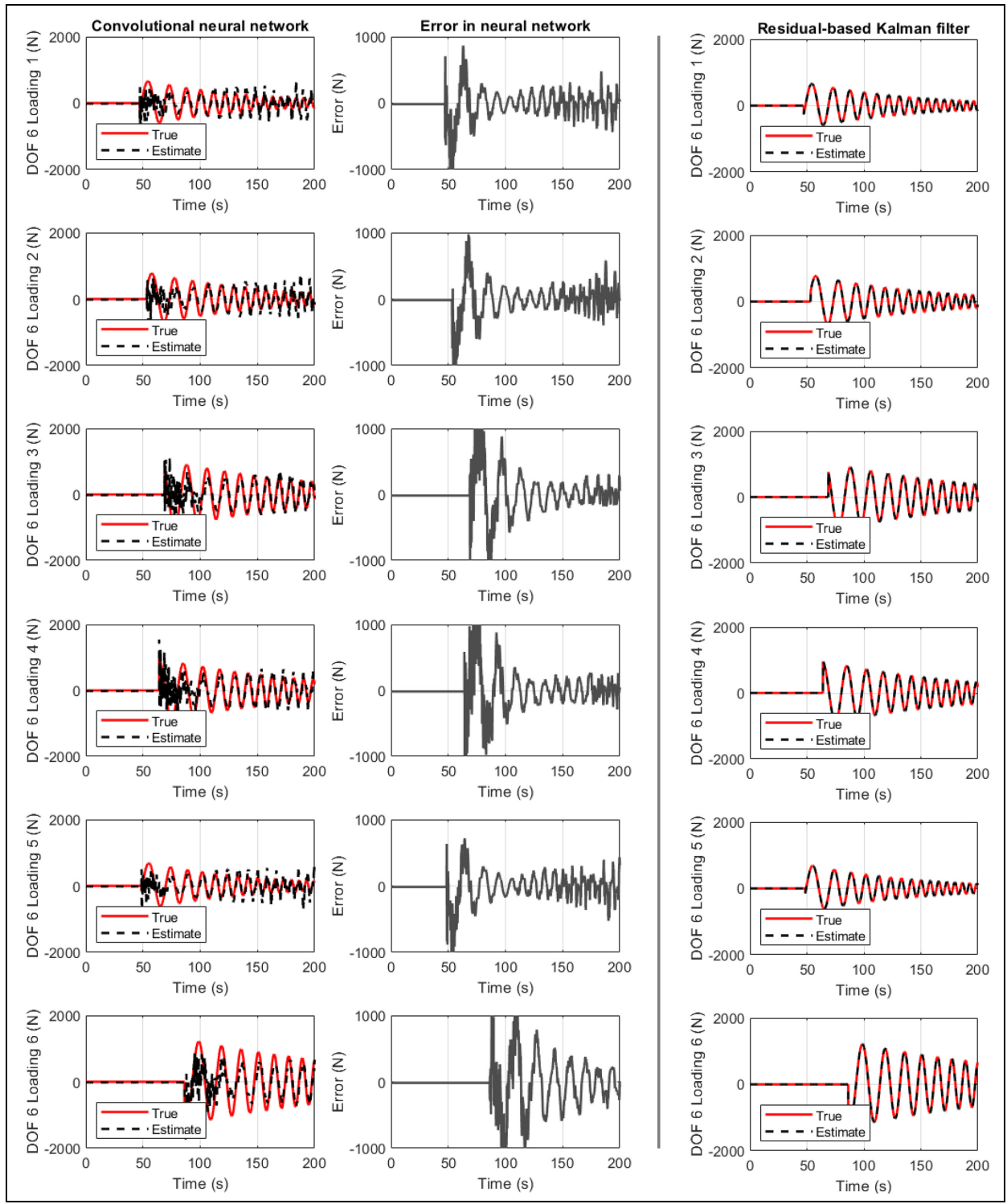

**Figure 5.** Structure of section "Structural loading identification in a 6-story building": results for the 6-story shear-type building when the convolutional neural network is used. First column: true and estimated loading at floor 6. Second column: error at loading identification. Third column: Residual-based Kalman filter performance.

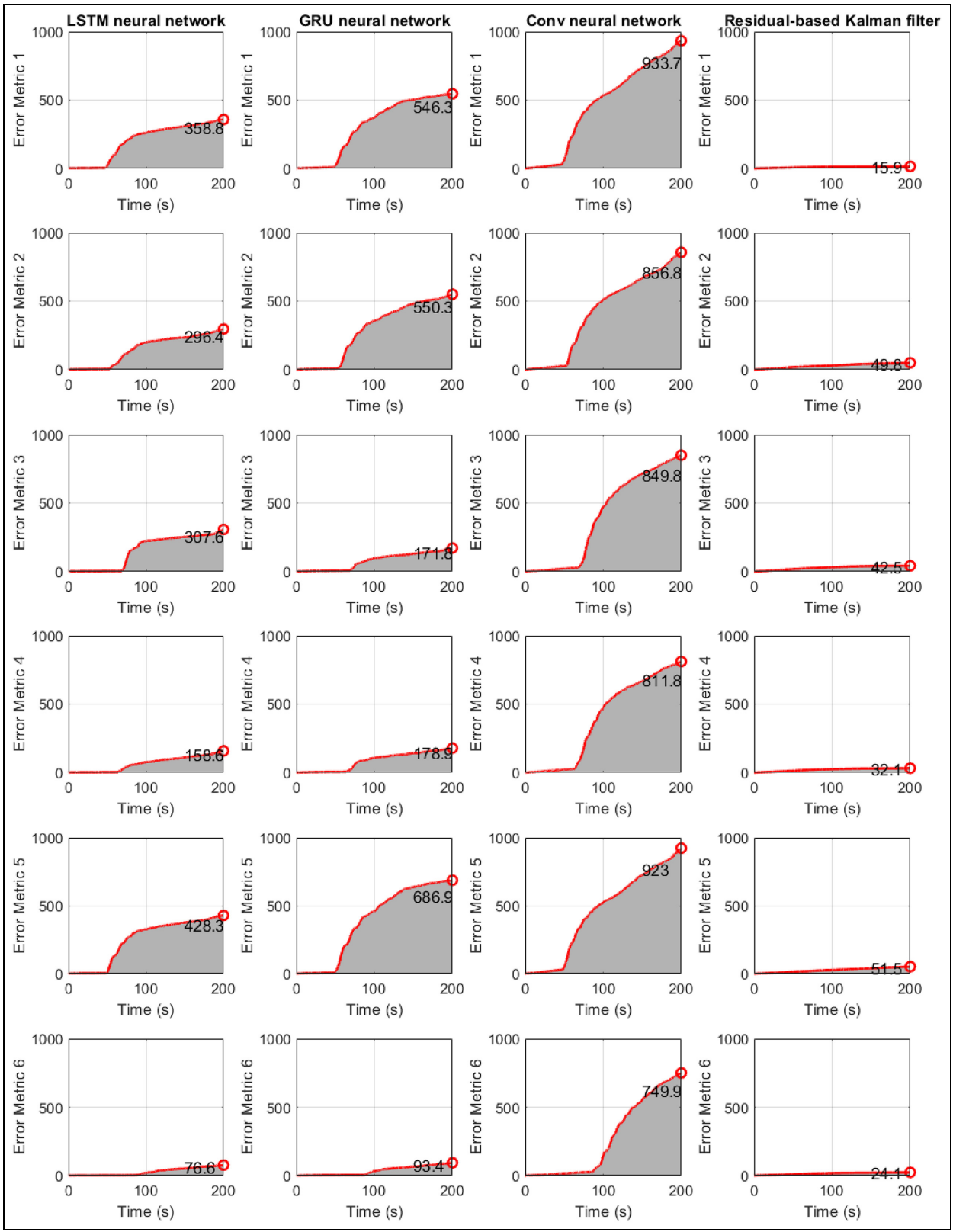

**Figure 6.** Structure of section "Structural loading identification in a 6-story building": results for the 6-story shear-type building when the error metric $E(t)$ of Equation (25) is used. First column: LSTM neural network. Second column: GRU neural network. Third column: CNN. Fourth column: Residual-based Kalman filtering.
LSTM: long short-term memory; GRU: gated recurrent unit; CNN: convolutional neural network.

**Table 1.** Final value of error metric $E(t)$ of Equation (25) for the 6-story building of section "Structural loading identification in a 6-story building."

| Case | LSTM network | GRU network | Conv network | Residual KF |
|---|---|---|---|---|
| ● DOF 6 Loading 1 | 358.8 | 546.3 | 933.7 | 15.9 |
| ● DOF 6 Loading 2 | 296.4 | 550.3 | 856.8 | 49.8 |
| ● DOF 6 Loading 3 | 307.6 | 171.8 | 849.8 | 42.5 |
| ● DOF 6 Loading 4 | 158.6 | 178.9 | 811.8 | 32.1 |
| ● DOF 6 Loading 5 | 428.3 | 686.9 | 923.0 | 51.5 |
| ● DOF 6 Loading 6 | 76.6 | 93.4 | 749.9 | 24.1 |

LSTM: long short-term memory; DOF: degree of freedom; GRU: gated recurrent unit; KF: Kalman filter.

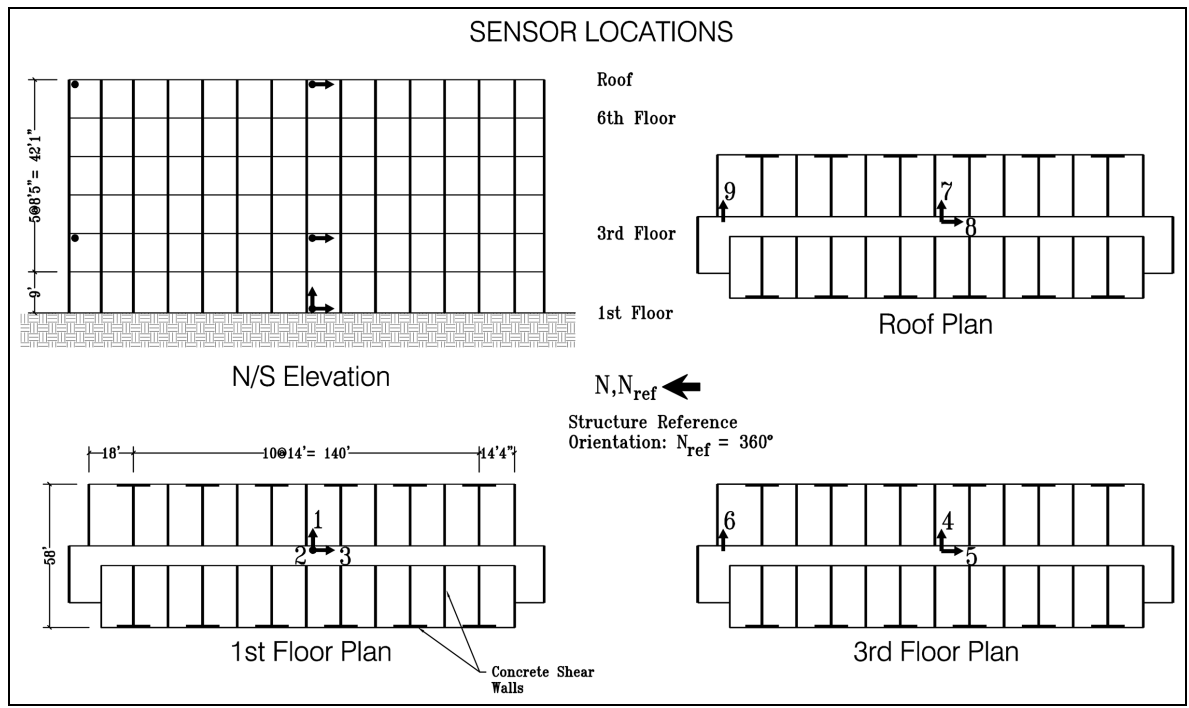

**Figure 7.** Sensor layout of the 6-story hotel in San Bernardino of section "Structural loading identification for a hotel in San Bernardino" (Station No. 23287).
Source: http://www.strongmotioncenter.org/

Figure 8 refers to the case where the LSTM neural network is used. The results are satisfactory for all six cases. On the other side, the LSTM neural network has the highest computation cost, and it is seen as the least favorable (Table 2).

Figure 9 refers to the case where the GRU neural network is used. The results are often more satisfactory for all six cases, and with a convergence time reduction compared to the LSTM network (Table 2).

Figure 10 refers to the case where the 1D CNN is used. The results are the most satisfactory for all six cases compared to the previous networks. Along these lines, a clear and significant convergence time reduction is observed (Table 2). It can be then concluded that for the base excitation, the 1D CNN identifies and predicts the loading with better accuracy and with a less computation time. This result is not true for the top floor excitation of the previous investigation of section "Structural loading identification in a 6-story building."

Finally, Figure 11 refers to the case where the error metric $E(t)$ of Equation (25) is used. The results show that the LSTM network has (relatively) poor performance with higher computation cost, shown in section "Discussion." In general, it seems that the performance of three neural networks is better compared to the building of section "Structural loading identification in a 6-story building." This implies that base excitation results in a better learning for the networks when the responses are all related to the input, than exciting only a single DOF as in section "Structural loading identification in a 6-story building."

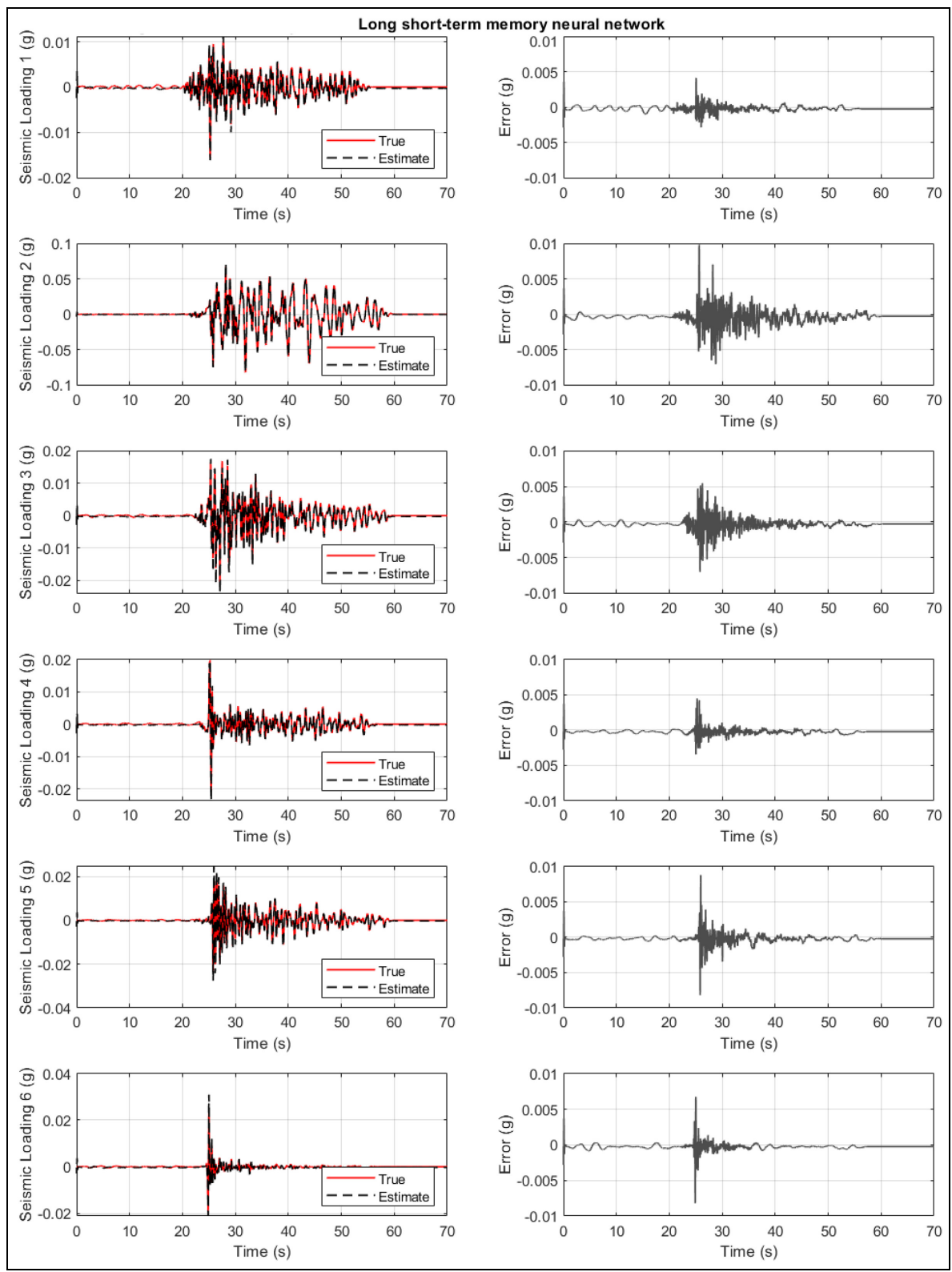

**Figure 8.** Structure of section "Structural loading identification for a hotel in San Bernardino": results for the 6-story hotel in San Bernardino when the long short-term memory neural network is used. First column: true and estimated loading at the base. Second column: error at loading identification.

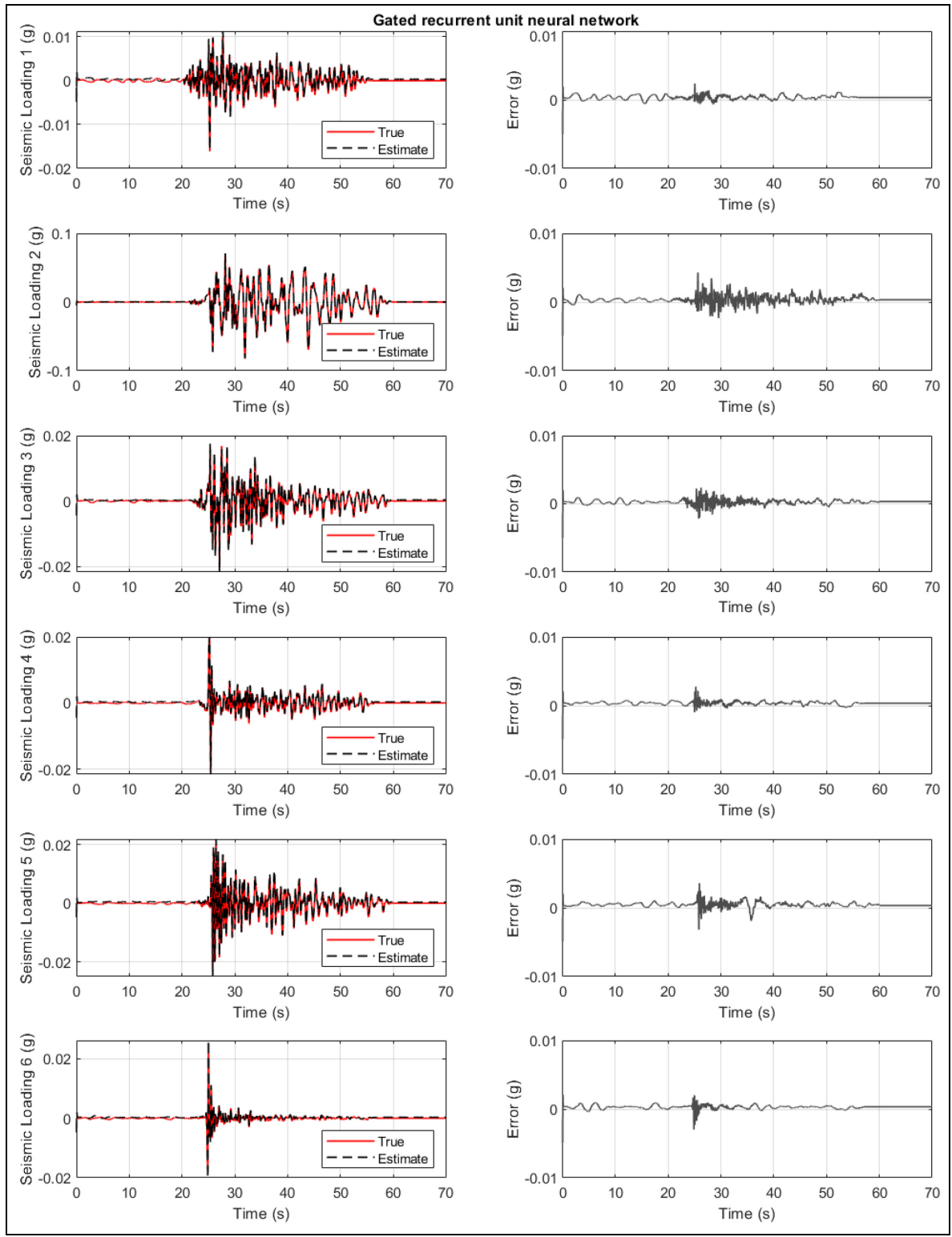

**Figure 9.** Structure of section "Structural loading identification for a hotel in San Bernardino": results for the 6-story hotel in San Bernardino when the gated recurrent unit neural network is used. First column: true and estimated loading at the base. Second column: error at loading identification.

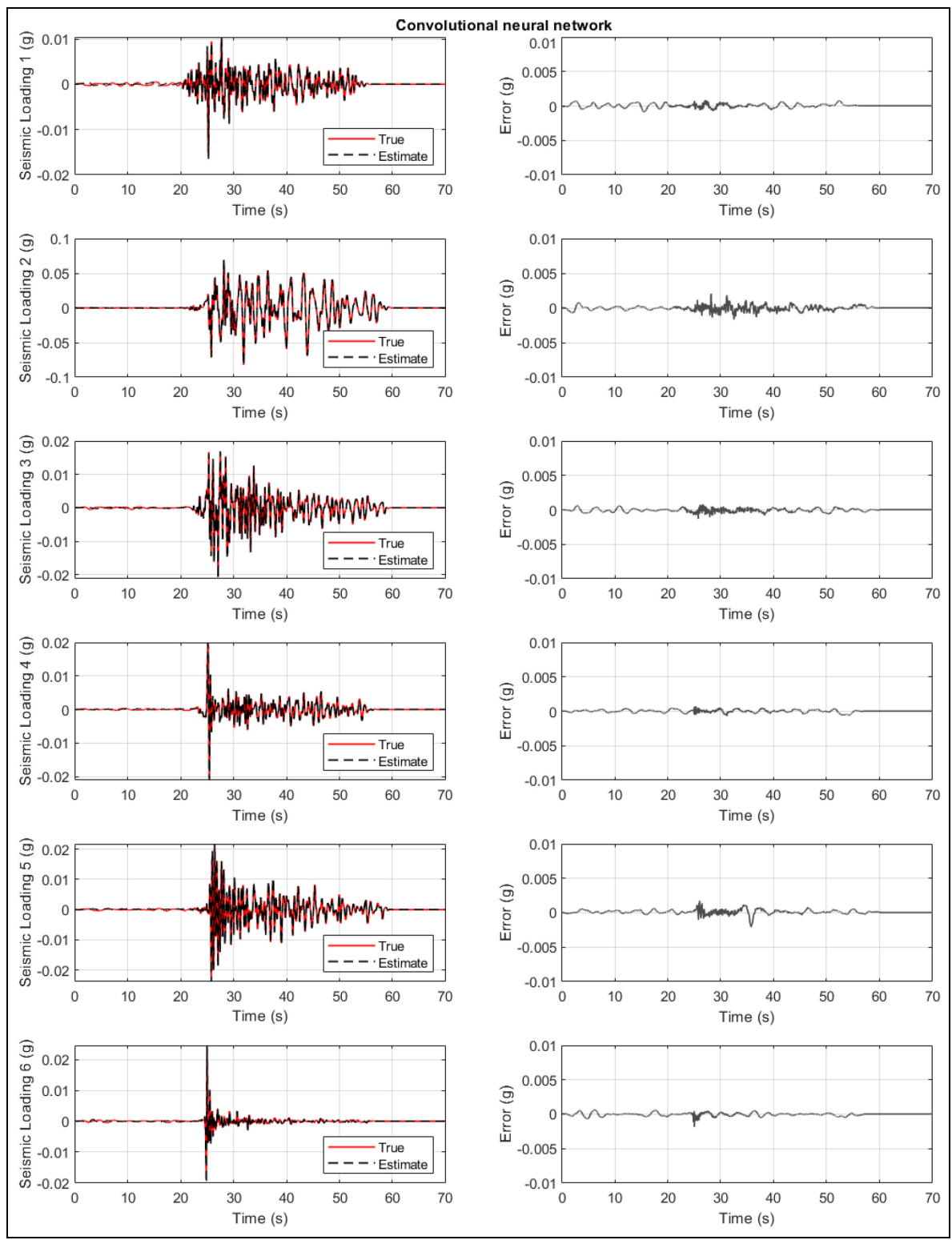

**Figure 10.** Structure of section "Structural loading identification for a hotel in San Bernardino": results for the 6-story hotel in San Bernardino when the convolutional neural network is used. First column: true and estimated loading at the base. Second column: error at loading identification.

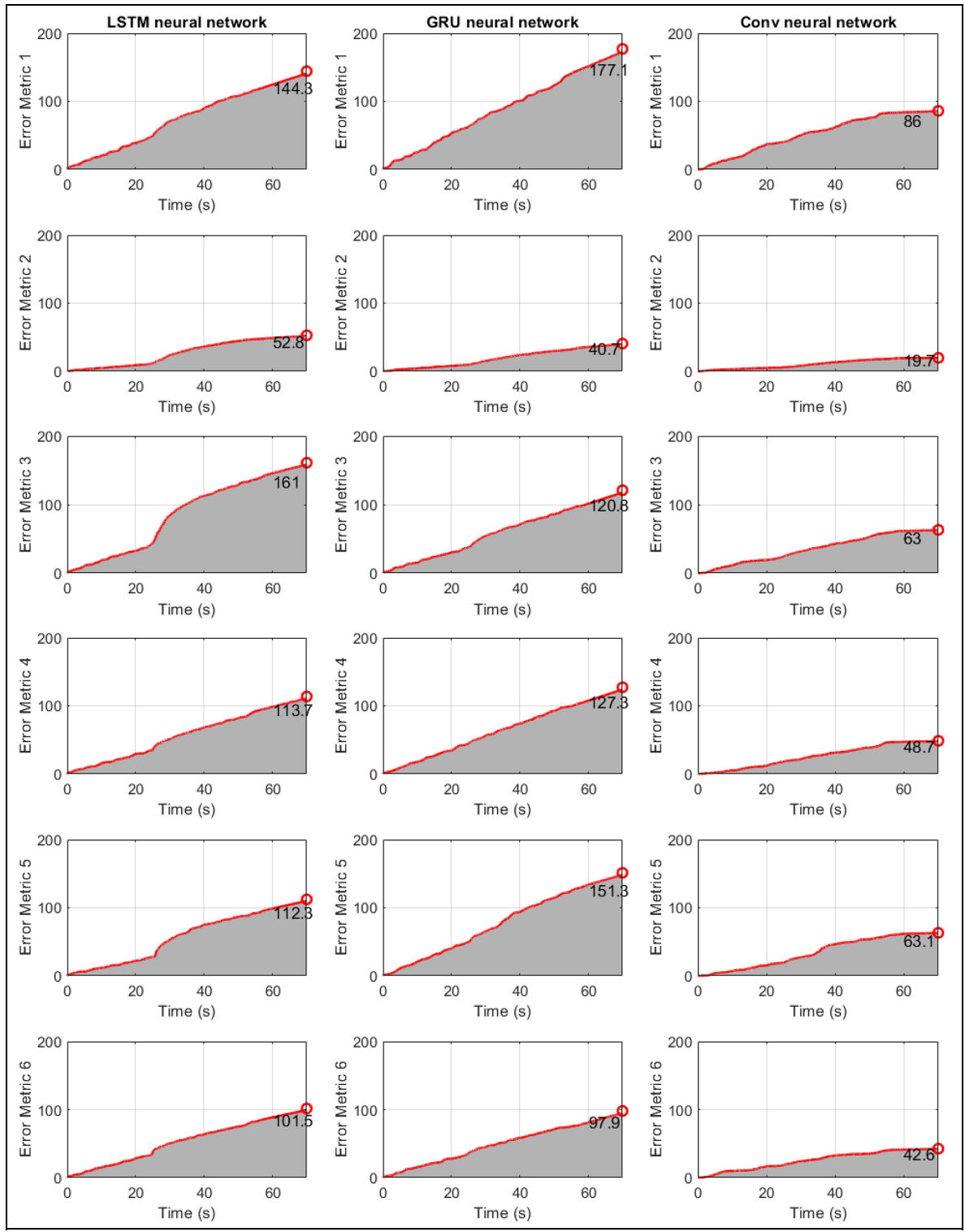

**Figure 11.** Structure of section "Structural loading identification for a hotel in San Bernardino": results for the 6-story hotel in San Bernardino when the error metric $E(t)$ of Equation (25) is used. First column: LSTM neural network. Second column: GRU neural network. Third column: CNN.
LSTM: long short-term memory; GRU: gated recurrent unit; CNN: convolutional neural network.

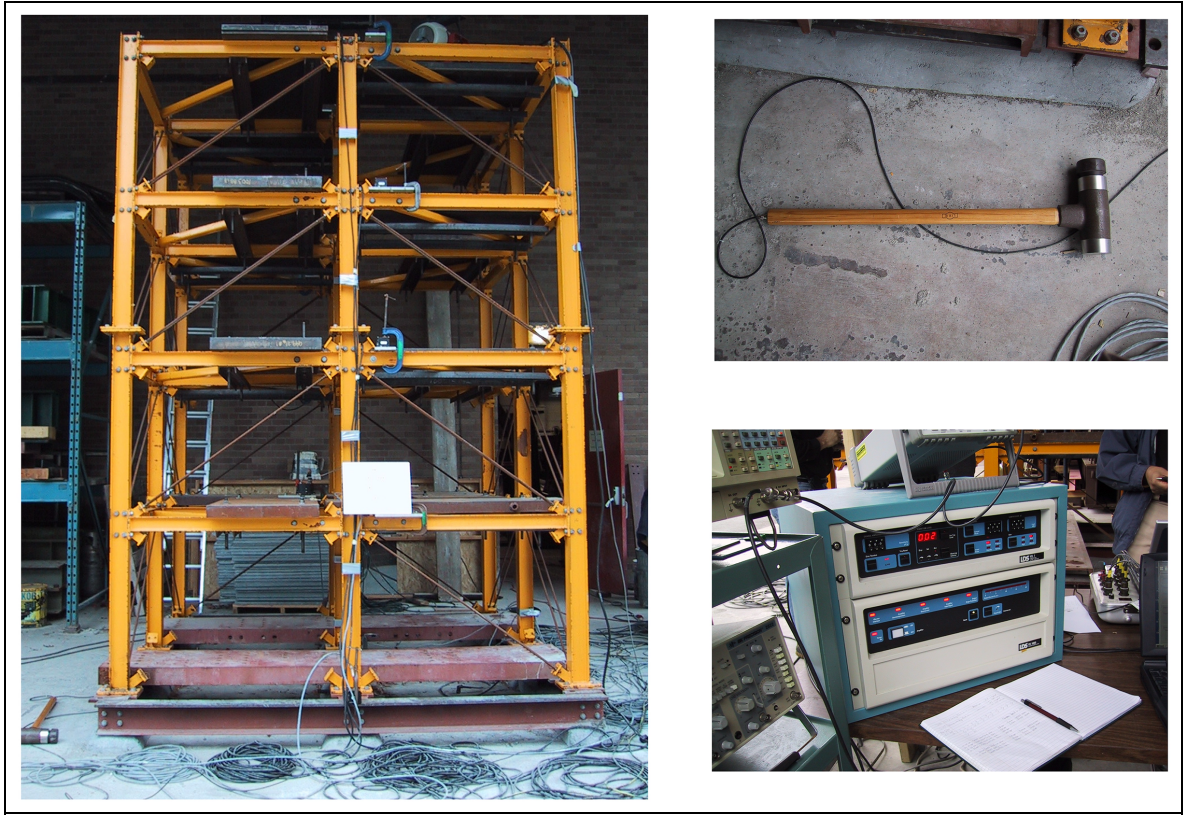

**Figure 12.** IASC-ASCE structural health monitoring benchmark building of section "Structural loading identification in the IASC-ASCE structural health monitoring benchmark problem", the Dynatron 5803A 12 lbf Impulse Hammer, and the monitor and console equipment.[115]

## Structural loading identification in the IASC-ASCE structural health monitoring benchmark problem

The proposed methodologies are also examined in a hammer-type loading scenario. This examination corresponds to the second phase of experiments conducted by the IASC-ASCE Structural Health Monitoring Task Group, tested at the University of British Columbia.[115–118] The study focuses on applying structural health monitoring techniques to data collected from a 4-story, 2-bay by 2-bay steel-frame structure, as shown in Figure 12. The structure, measuring 2.5 × 2.5 m in plan and 3.6 m tall, is mounted on a concrete slab outside the testing laboratory. To enhance realism, mass distribution was involved placing floor slabs in each bay per floor, with off-center masses on each floor.[115] The experimental setup included three types of excitation: electrodynamic shaker, impact hammer, and ambient vibration. Accelerometers strategically placed across the structure facilitated the measurement of structural responses.

Fifteen accelerometers were positioned throughout the frame and the base to capture the responses of the test structure. The placement included sensors for measuring northsouth and eastwest motion.

The excitation and impact hammer tests employed a Dynatron 5803A 12 lbf Impulse Hammer. This hammer, equipped with a force transducer, recorded measurements during tests involving 3–5 hits. Impact locations were chosen on the south and east faces of the first floor in the southeast corner. A force transducer on the hammer tip measured the force input during impact tests. A 16-channel DasyLab acquisition system recorded structural responses, with sampling rates of 250 Hz for shaker and ambient tests, and 1000 Hz for hammer tests. Anti-aliasing filters were applied selectively, and the data acquisition system commenced before the first impact, recording a series of hits within each test.

Since there was a very limited amount of data for the same damage scenario, namely the same structure,

the signals of multiple hammer impact split into smaller (four) signals of a single hammer impact in them. Finally, the networks are trained with two signals, validated with one signal, and finally tested on a final signal. This approach examines the capability of the networks on extremely limited datasets. Importantly, the neural network architectures are defined similarly to section "Structural loading identification in a 6-story building."

All three network cases are examined on Figure 13. They show the true and the identified structural load (first column) for the unknown load where the network never trained or validated, and the load identification error on the second figure column. In all cases, acceleration measurement are selected from all stories.

Figure 13 top plots refer to the case where the CNN is used. The results are the most satisfactory compared to the other two models. At the same time, the CNN has the lowest computation cost, and it is seen as the most favorable. The error seen is attributed to the delay on the impact load time, and not to the wrong amplitude.

Figure 13 middle plots refer to the case where the LSTM neural network is used. The results are not satisfactory and with an additional convergence time compared to the previous case.

Figure 13 bottom plot refers to the case where the GRU neural network is used. The results are also not satisfactory, but with a shorter convergence time compared to the previous case.

Finally, Figure 14 refers to the case where the error metric $E(t)$ of Equation (25) is used. The results show that the convolutional network performs better and with lower computation cost, shown in section "Discussion."

The poor performance on this investigation of the LSTM and GRU neural network is expected. The intuition behind them is to create an additional module in a neural network that learns when to remember and when to forget some characteristic of the provided signal. In other words, the network, effectively learns which patterns might be needed in the signal and when that information is no longer needed. This poses a disadvantage for structural load identification in a hammer impact case as it seen as an unexpected excitation in the structure. This is wrongly assumed to not be attributed to structure response or play an important role in the final prediction, and therefore it is neglected. In the hammer test scenario, this "unexpected" excitation is correct and the networks wrongly forget and neglect it.

## Discussion

The presented work provided a simple, yet effective, way to identify the load in structural dynamics. It did not aim to present a machine learning algorithm advancement, rather than to apply the vast capabilities of such tools to the structural load identification problem. To this end, the efficiency and robustness of the methods were tested to both simulated and real data, and in different loading types.

This work provided an assessment for the GRU networks, LSTM networks, and CNN in the framework of limited datasets. For the structural health monitoring case of civil structures, this is realistic. All the presented tools can perform much better in a big data availability scenario, but this is often impractical. Despite the small dataset investigation, all the tools shown a great capability.

Regarding the network algorithm parameters, the examinations so far showed a recommendation of high values for the filter size and the number of neurons in the layers. The first one defines the kernel where the data are multiplied by, while the second one determines the number of feature maps. However, for the case of the dropout layer parameter, using a large number may lead to a poorer performance. This is illustrated in Figure 15, where the system of section "Structural loading identification for a hotel in San Bernardino" under seismic loading was modeled using the dropout layer value of 0.75.

The recommendation of high values for the filter size and the number of neurons in the layers sounds restrictive or suboptimal since it leads to higher weights for back-propagation, and ultimately to higher computational cost. Despite this, the computational cost of this approach is bearable. This is attributed to two main reasons: the one-dimensional nature of the data, and the small dataset training approach which was implemented. Future research is recommenced on the optimal value of them, or improved network architectures. The author investigated improving further the computational cost by removing layers from the architecture of section "Structural loading identification in a 6-story building." Specifically a set of GRU, ReLu, and dropout layer is removed; however, this resulted in faster convergence but with a poorer performance (seen in Figure 16).

Regarding the concern about the robustness of proposed approaches against the noise effect, simulation are provided. Here, the data are contaminated by a Gaussian white noise sequence with a 10%, 15%, and 20% root-mean-square noise-to-signal ratio for the Kalman filter, and 15% root-mean-square noise-to-signal ratio for the neural networks. The dynamic load identification accumulated error $E(t)$ of Equation (25) is shown for all noise levels of the Kalman filter in Figure 17. The higher noise level results is higher error.

The presented Kalman filter approach performs joint input-state-parameter estimation. The results for parameter estimation in the section "Structural loading identification in a 6-story building" building study are also shown in Figure 18 for the fourth floor parameter and for all noise levels. More parameter and noise results are shown in Impraimakis and Smyth.[41] The parameter estimation slowly convergences to the true values for 10% noise, while for the higher noise cases, it takes even more time to convergence to the true values. The convergence may not be occur during the identification duration of 200 s in high-noise levels.

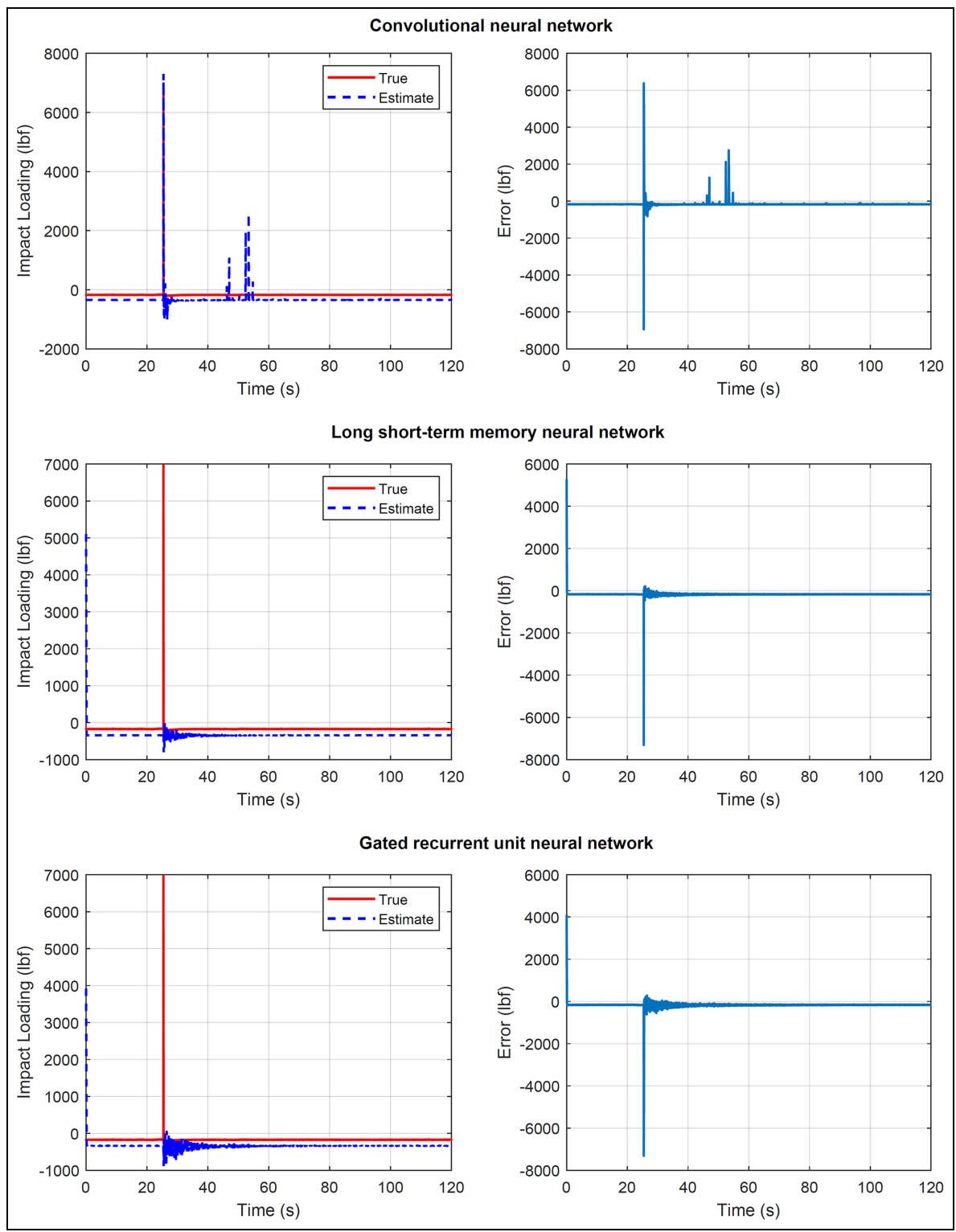

**Figure 13.** Structure of section "Structural loading identification in the IASC-ASCE structural health monitoring benchmark problem": results for the IASC-ASCE structural health monitoring benchmark problem when the LSTM, GRU, and CNN are used. First column: true and estimated loading on the impact hammer scenario. Second column: error at loading identification. LSTM: long short-term memory; GRU: gated recurrent unit; CNN: convolutional neural network.

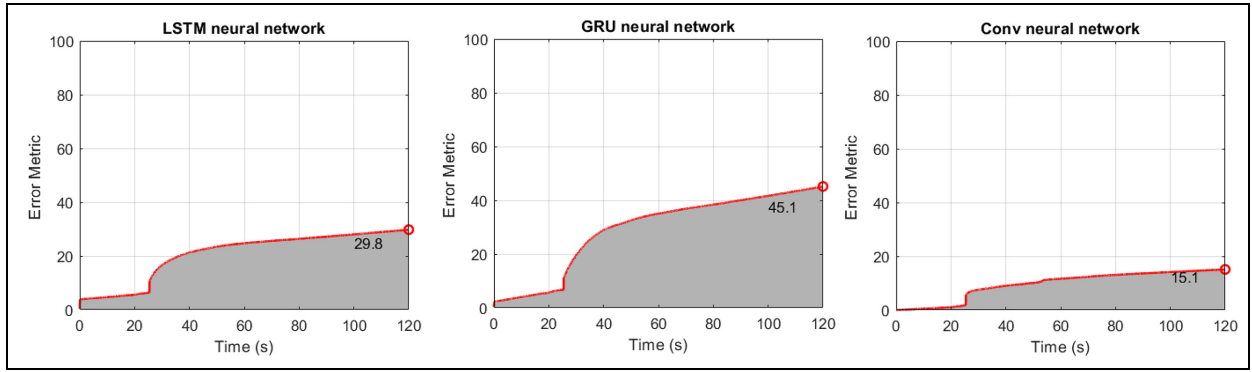

**Figure 14.** Structure of section "Structural loading identification in the IASC-ASCE structural health monitoring benchmark problem": Results for the IASC-ASCE structural health monitoring benchmark problem when the error metric $E(t)$ of Equation (25) is used.

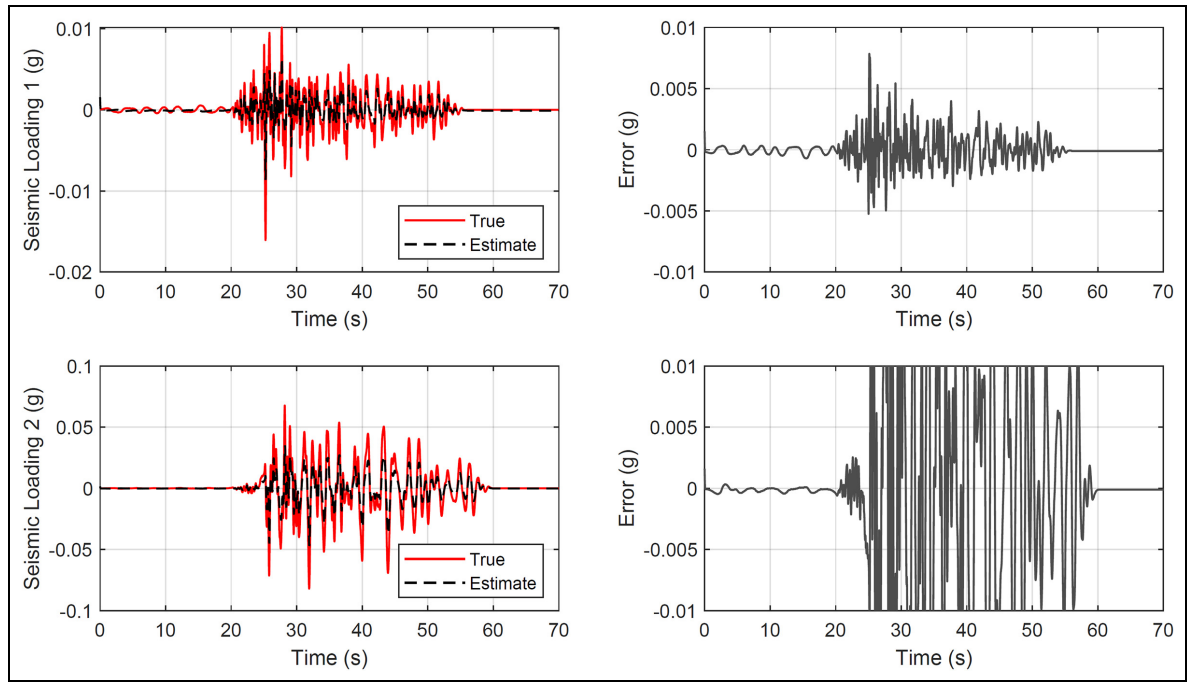

**Figure 15.** Structure of section "Structural loading identification for a hotel in San Bernardino" in section "Discussion": results for the 6-story hotel in San Bernardino when the gated recurrent unit neural network is used with dropout layer value of 0.75. First column: true and estimated loading at the base. Second column: error at loading identification.

An investigation is also made for the networks when the data are contaminated by a Gaussian white noise sequence with a 15% root-mean-square noise-to-signal ratio. All networks underperformed compare to section "Structural loading identification in a 6-story building," where the high noise is inserted to the prediction. Figure 19 shows the dynamic load identification for the first case of each network, the error compared to true loading, and the accumulated error of Equation (25).

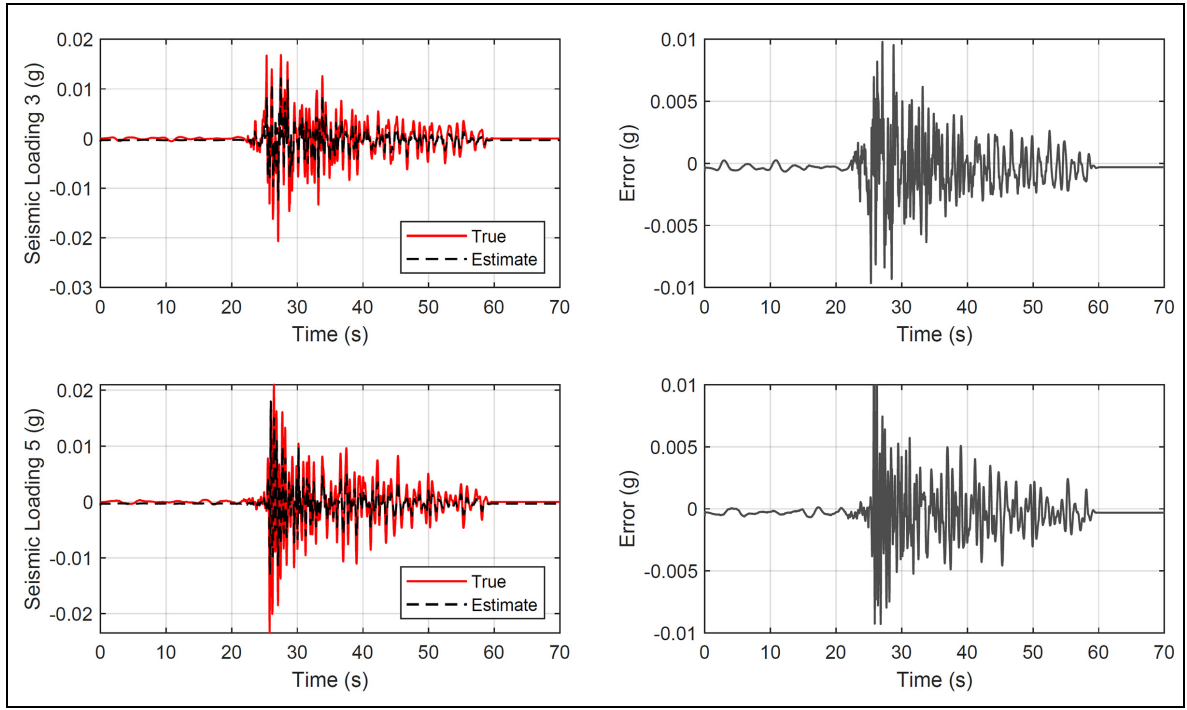

**Figure 16.** Structure of section "Structural loading identification for a hotel in San Bernardino" in section "Discussion": Results for the 6-story hotel in San Bernardino when the gated recurrent unit neural network is used with less network layers. First column: true and estimated loading at the base. Second column: error at loading identification.

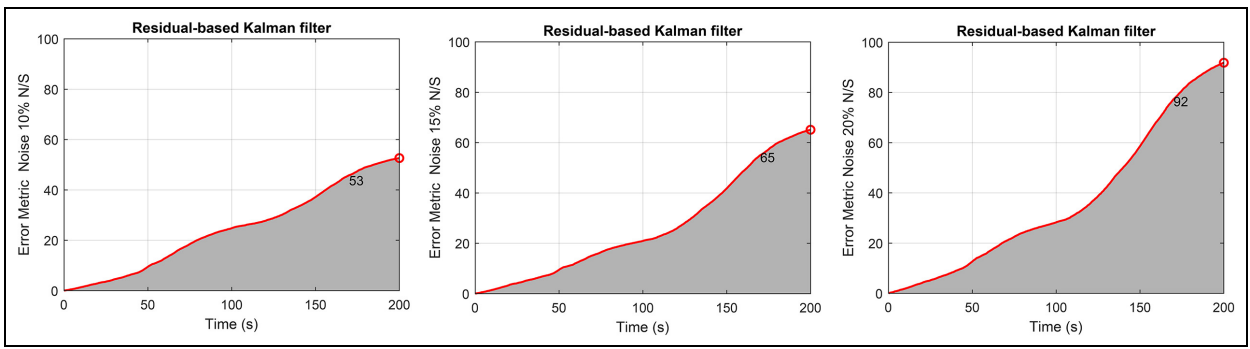

**Figure 17.** Structure of section "Structural loading identification in a 6-story building" in section "Discussion": results for the 6-story shear-type building when the data are contaminated by a Gaussian white noise sequence with a 10%, 15%, and 20% root-mean-square noise-to-signal ratio for the residual-based Kalman filter. Accumulated error of equation for 10% noise (left plot), 15% noise (middle plot), and 20% noise (right plot).

Finally, Figure 20 shows the Kalman filter approach for the impact load identification case for the section "Structural loading identification in a 6-story building" building case. It is not applied directly to the IASC-ASCE structural health monitoring benchmark problem to avoid first creating a nonphysically parametrized reduced order model; a task suggested for future research.

Another concern is related to the data-driven only training of the presented tools. It has observed[110,119–126] that by including a physics-aware constraint or a mathematical model, the training is improved. However, this is not always practical for large-scale structures as it requires full system identification, which finally results in the need of even greater data collection. The tools presented here do not require any parameter estimation in order to perform the structural load identification. It is important to mention that in the case where a physics-based model is available, the computational cost of the training is reduced for all neural networks,

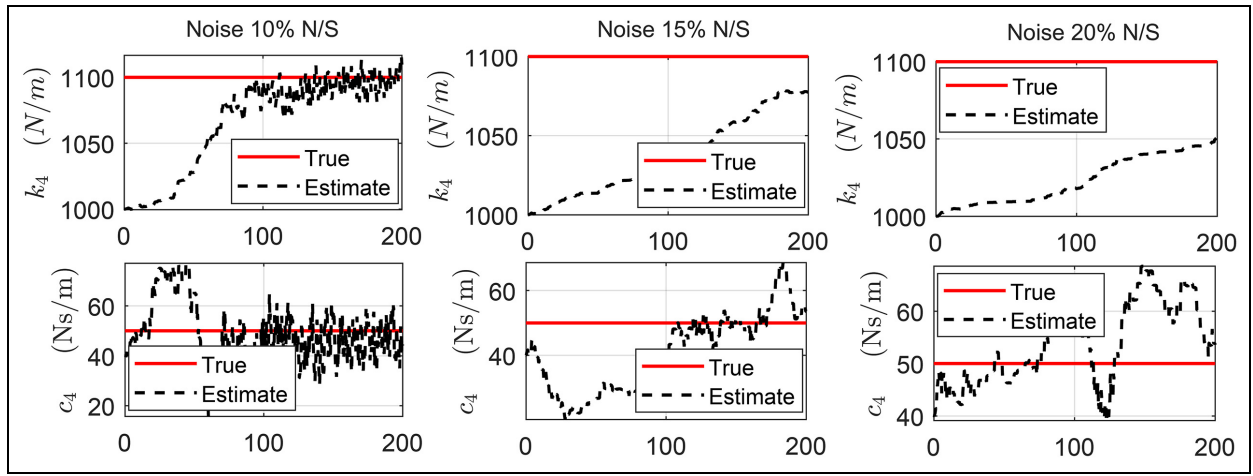

**Figure 18.** Structure of section "Structural loading identification in a 6-story building" in section "Discussion": Results for the 6-story shear-type building when the data are contaminated by a Gaussian white noise sequence with a 10%, 15%, and 20% root-mean-square noise-to-signal ratio for the residual-based Kalman filter. True and estimated stiffness and damping DOF four parameters for 10% noise (first column), 15% noise (second column), and 20% noise (third column).
DOF: degree of freedom.

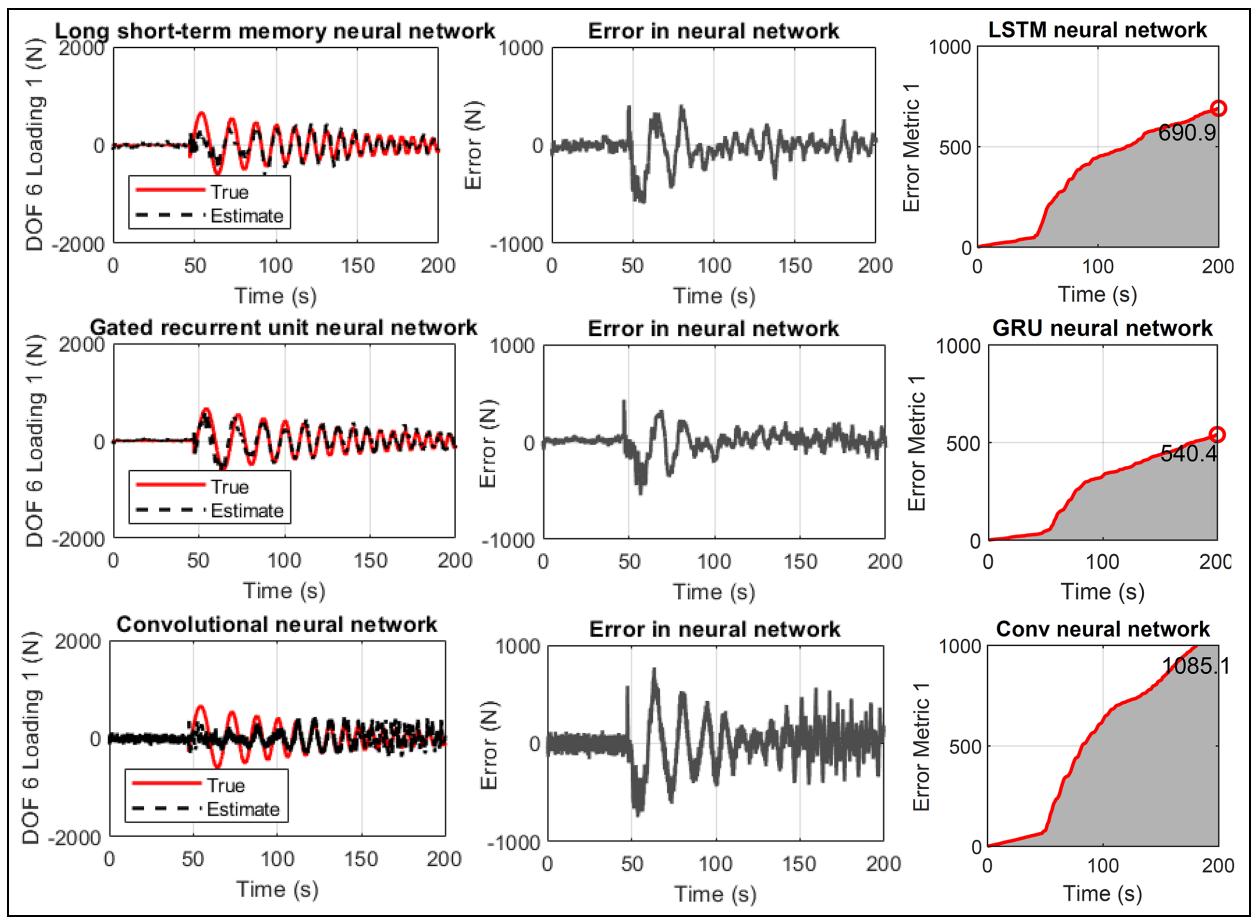

**Figure 19.** Structure of section "Structural loading identification for a hotel in San Bernardino" in section "Discussion": Results for the 6-story hotel in San Bernardino when the data are contaminated by a Gaussian white noise sequence with a 15% root-mean-square noise-to-signal ratio for all networks. True and estimated loading, error compared to true loading, and accumulated error of Equation (25) for the long short-term memory network (first row), the gated recurrent network (second row), and the convolutional network (third row).

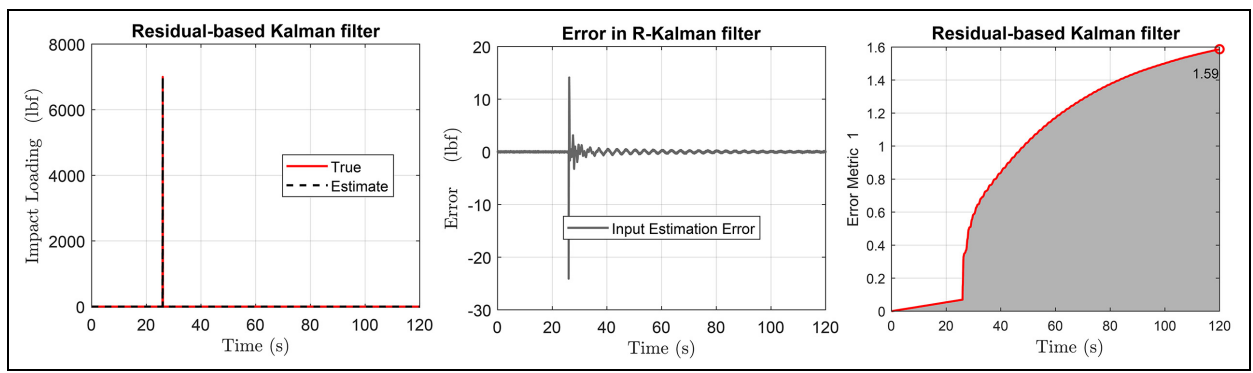

**Figure 20.** Structure of section "Structural loading identification in a 6-story building" in section "Discussion": results for the 6-story shear-type building when the impact load of section "Structural loading identification in the IASC-ASCE structural health monitoring benchmark problem" is examined with the residual-based Kalman filter. True and estimated loading (first plot), error compared to true loading (middle plot), and accumulated error of Equation (25) (last plot).

**Table 2.** Computational time for training the presented networks in minutes per 1000 epochs.

| Case | LSTM network | GRU network | Conv network |
|---|---|---|---|
| ● Shaker loading of section "Structural loading identification in a 6-story building" | 337.13 min/1000e | 310.14 min/1000e | 7.44 min/1000e |
| ● Seismic excitation loading of section "Structural loading identification for a hotel in San Bernardino" | 461.31 min/1000e | 452.95 min/1000e | 15.57 min/1000e |
| ● Hammer loading of section "Structural loading identification in the IASC-ASCE structural health monitoring benchmark problem" | 1058.69 min/1000e | 1044.82 min/1000e | 7.21 min/1000e |

LSTM: long short-term memory; GRU: gated recurrent unit.

and the accuracy is improved. Specifically for the training time, it was shown in this study that the CNN has the lowest one, while between the rest two, GRU has the shorter convergence time. Table 2 shows the computational time for training the presented networks in minutes per 1000 epochs for all networks and applications. The computer used has processor 12th Gen Intel(R) Core(TM) i7-12850HX (24 CPUs), 2 NVIDIA RTX A2000 8 GB GPUs, and RAM 32.0 GB.

## Future research

Relating to the joint input-state estimation, the methodologies require knowledge of the structural model and parameters. This may be infeasible, or it may require the collection of additional data to perform full system identification. A way to address this issue is by the use of the joint input-parameter-state estimation methodologies such as the RKF of section "Dynamic load identification using physics-based residual Kalman filtering." However, these methodologies also have main deficiencies. For instance, the nature of the loading has to be of zero mean value to be filtered, or the requirement of having a known location of the loading, or known zero-values inputs at known location for identifiability reasons.[113,127,128] Contrastingly, for a different combination or number of measurements, different convergence timing is observed for all networks, but unidentifiability issues are not occurred.

By performing a data-driven only approach, the user also does not have to consider different model classes and the select the optimal one. Those approaches calculate the evidence of each candidate model given the available measured data, and they finally select the simpler ones over the unnecessarily complicated ones. The importance of those methods is highlighted by the fact that a more complicated model fits the data better than one which has fewer adjustable uncertain parameters. This is attributed to the parameter fitting which depends too much on the detail of the data and the measurement noise. On the other hand, the presented networks solve the structural load identification problem without a need to select the structural model class.

Another concern is related to the investigation of different structures than buildings. In reality, in another case such as in a bridge investigation, the loading may not be directly sensitive to all responses. As a result, the networks could perform poorly. Additional research is therefore suggested for civil structures different than buildings.

Another concern is related to the investigation into the extrapolation capabilities of the approach since only the inputs–outputs are used for the training and the load identification. The examinations so far showed the potential of the method when the structural model remains the same. However, this assumption may not be true if a change happen to the structure, some damage for instance, or any other modification on the structure. The author slightly changed the simulated structural model of section "Structural loading identification in a 6-story building," keeping the same trained neural network models, and they all underperformed. This does not occur in the physics-based Kalman filter approach. As a result, the deep learning approaches are not capable of some form of extrapolation to predict structural load for structures with properties outside of the training dataset to ensure good performance. When employed on a real engineering system where the structure may change, one must have some prior belief about the expected model patterns in order to generate comprehensive training datasets. This will lead to retrain the network for future good prediction. This is a pertinent test for structural load approaches in engineering applications as there could be high-cost and safety critical ramifications if the loading is confidently predicted incorrectly.

Regarding applying the Kalman filter approach for input estimation in the case study of base-excited building section "Structural loading identification for a hotel in San Bernardino," the current work did not assume the extra information of known model parameter in Kalman filtering for a fair comparison with the network in the same dataset. This case obviously results in an even better performance of Kalman filtering presented already by Eftekhar Azam et al.[114] A limitation and future suggestion is then how to implement the residual-based Kalman filtering for scenario of base excitation (which excites all DOFs) where input-parameter-state estimation fails the identifiability tests.[113] Similarly, for the case of section "Structural loading identification in the IASC-ASCE structural health monitoring benchmark problem," it requires a reduced order modeling which results in a nonphysical parameter estimation not examined here, and it is suggested for future research. A future suggestion lies also in combing Kalman filtering and neural networks as exists for dynamic state estimation.[3]

A final concern is related to the uncertainty quantification where the structural load identification methodology should provide.[129–131] This is a desirable property for the structural load prediction approaches to possess that accurately representing the uncertainty around predictions. In the framework of GRU, LSTM, and CNN, this may be crudely achieved by retraining the model multiply times and take the average and the rest statistical properties of the network prediction, or by using a variational inference approach, while for the Kalman filter by incorporating the unknown input in the state vector.[23]

## Conclusion

The dynamic structural load identification capabilities of the GRU, LSTM, and CNNs were examined herein. The examination was on realistic small dataset training conditions, and on a comparative view to the physics-based RKF. The dynamic load identification suffers from the uncertainty related to obtaining poor predictions when in civil engineering applications only a low number of tests are performed or are available, or when the structural model is unidentifiable. In considering the methods, first, a simulated structure was investigated under a shaker excitation at the top floor. Second, a building in California was investigated under seismic base excitation, which results in loading for all DOFs. Finally, the IASC-ASCE structural health monitoring benchmark problem was examined for impact and instant loading conditions.

Overall, these network methods allowed for structural load identification with

1. No need for data filtering for reasonable noise levels.
2. No need for system identification, known structural parameters, or a structural model.
3. Real-time prediction when the networks are trained.
4. Capability of providing the structural load identification for all loading types, with respect to the use of the appropriate network each time.
5. Reasonable computational cost for small datasets scenarios.

Importantly, the methods were shown to outperform each other on different loading scenarios, while the RKF was shown to outperform the networks in physically parametrized identifiable cases.

## Acknowledgements

The author would like to gratefully acknowledge the reviewers for their constructive comments, Editor T.C. for the friendly communication, Andrew W. Smyth for the previous insightful discussions on residual-based Kalman filtering, and the Center for Engineering Strong Motion Data and the Structural Health Monitoring Task Group for providing the data.

## Declaration of conflicting interests

## Funding

The author received no financial support for the research, authorship, and/or publication of this article.

## ORCID iD

Marios Impraimakis  https://orcid.org/0000-0001-9350-2680

## References

1. Liu R, Dobriban E, Hou Z, et al. Dynamic load identification for mechanical systems: a review. *Arch Comput Methods Eng* 2022; 29(2): 831–863.
2. Ching J, Beck JL, Porter KA, et al. Bayesian state estimation method for nonlinear systems and its application to recorded seismic response. *J Eng Mech* 2006; 132(4): 396–410.
3. Liu W, Lai Z, Bacsa K, et al. Neural extended Kalman filters for learning and predicting dynamics of structural systems. *Struct Health Monit* 2024; 23(2): 1037–1052.
4. Mansouri M, Avci O, Nounou H, et al. A comparative assessment of nonlinear state estimation methods for structural health monitoring. In: *Model validation and uncertainty quantification, Volume 3: Proceedings of the 33rd IMAC, a conference and exposition on structural dynamics*, Cham, 2015, pp. 45–54. Springer.
5. Wu R-T and Jahanshahi MR. Deep convolutional neural network for structural dynamic response estimation and system identification. *J Eng Mech* 2019; 145(1): 04018125.
6. Zhang R, Chen Z, Chen S, et al. Deep long short-term memory networks for nonlinear structural seismic response prediction. *Comput Struct* 2019; 220: 55–68.
7. Oh BK, Glisic B, Kim Y, et al. Convolutional neural network-based wind-induced response estimation model for tall buildings. *Comput Aided Civ Infrastruct Eng* 2019; 34(10): 843–858.
8. Guarize R, Matos NAF, Sagrilo LVS, et al. Neural networks in the dynamic response analysis of slender marine structures. *Appl Ocean Res* 2007; 29(4): 191–198.
9. Ying W, Chong W, Hui L, et al. Artificial neural network prediction for seismic response of bridge structure. In: *2009 International conference on artificial intelligence and computational intelligence*, Shanghai, China, 2009, vol. 2, pp. 503–506. IEEE.
10. Papadrakakis M, Papadopoulos V and Lagaros ND. Structural reliability analyis of elastic-plastic structures using neural networks and monte carlo simulation. *Comput Methods Appl Mech Eng* 1996; 136(1–2): 145–163.
11. Christiansen NH, Høgsberg J and Winther O. Artificial neural networks for nonlinear dynamic response simulation in mechanical systems. *Nordic Seminar Comput Mech* 2011; 24: 77–80.
12. Masri SF, Smyth AW, Chassiakos AG, et al. Application of neural networks for detection of changes in nonlinear systems. *J Eng Mech* 2000; 126(7): 666–676.
13. Olivier A and Smyth AW. A marginalized unscented Kalman filter for efficient parameter estimation with applications to finite element models. *Comput Methods Appl Mech Eng* 2018; 339: 615–643.
14. Impraimakis M and Smyth AW. Integration, identification, and assessment of generalized damped systems using an online algorithm. *J Sound Vibr* 2022; 523: 116696.
15. Ebrahimian H, Astroza R and Conte JP. Extended Kalman filter for material parameter estimation in nonlinear structural finite element models using direct differentiation method. *Earthquake Eng Struct Dynamics* 2015; 44(10): 1495–1522.
16. Chatzi EN, Hiriyur B, Waisman H, et al. Experimental application and enhancement of the XFEM–GA algorithm for the detection of flaws in structures. *Comput Struct* 2011; 89(7–8): 556–570.
17. Azam SE, Chatzi E and Papadimitriou C. A dual Kalman filter approach for state estimation via output-only acceleration measurements. *Mech Syst Signal Process* 2015; 60: 866–886.
18. Nayek R, Chakraborty S and Narasimhan S. A Gaussian process latent force model for joint input-state estimation in linear structural systems. *Mech Syst Signal Process* 2019; 128: 497–530.
19. Anagnostou G and Pal BC. Derivative-free Kalman filtering based approaches to dynamic state estimation for power systems with unknown inputs. *IEEE Trans Power Syst* 2017; 33(1): 116–130.
20. Liu Y, Wang L and Li M. Kalman filter–random forest-based method of dynamic load identification for structures with interval uncertainties. *Struct Control Health Monit* 2022; 29(5): e2935.
21. Law SS, Yung TH and Yuan XR. An interpretive method for moving force identification. *J Sound Vibr* 1999; 219(3): 503–524.
22. Chan THT. Recent research on identification of moving loads on bridges. *J Sound Vibr* 2007; 305(1–2): 3–21.
23. Lourens E, Reynders E, De Roeck G, et al. An augmented Kalman filter for force identification in structural dynamics. *Mech Syst Signal Process* 2012; 27: 446–460.
24. Ghahremani E and Kamwa I. Dynamic state estimation in power system by applying the extended Kalman filter with unknown inputs to phasor measurements. *IEEE Trans Power Syst* 2011; 26(4): 2556–2566.
25. Vettori S, Lorenzo ED, Peeters B, et al. An adaptive-noise augmented Kalman filter approach for input-state estimation in structural dynamics. *Mech Syst Signal Process* 2023; 184: 109654.
26. Hassanabadi ME, Liu Z, Azam SE, et al. A linear Bayesian filter for input and state estimation of structural systems. *Comput Aided Civ Infrastruct Eng* 2023; 38: 1749–1766.
27. Ji C-C and Liang C. A study on an estimation method for applied force on the rod. *Comput Methods Appl Mech Eng* 2000; 190(8–10): 1209–1220.
28. Valikhani M and Younesian D. Bayesian framework for simultaneous input/state estimation in structural and

mechanical systems. *Struct Control Health Monit* 2019; 26(9): e2379.

29. Naets F, Croes J and Desmet W. An online coupled state/input/parameter estimation approach for structural dynamics. *Comput Methods Appl Mech Eng* 2015; 283: 1167–1188.

30. Dertimanis VK, Chatzi EN, Eftekhar Azam S, et al. Input-state-parameter estimation of structural systems from limited output information. *Mech Syst Signal Process* 2019; 126: 711–746.

31. Ghorbani E, Buyukozturk O and Cha Y-J. Hybrid output-only structural system identification using random decrement and Kalman filter. *Mech Syst Signal Process* 2020; 144: 106977.

32. Castiglione J, Astroza R, Azam SE, et al. Auto-regressive model based input and parameter estimation for nonlinear finite element models. *Mech Syst Signal Process* 2020; 143: 106779.

33. Maes K, Karlsson F and Lombaert G. Tracking of inputs, states and parameters of linear structural dynamic systems. *Mech Syst Signal Process* 2019; 130: 755–775.

34. Lei Y, Xia D, Erazo K, et al. A novel unscented Kalman filter for recursive state-input-system identification of nonlinear systems. *Mech Syst Signal Process* 2019; 127: 120–135.

35. Song W. Generalized minimum variance unbiased joint input-state estimation and its unscented scheme for dynamic systems with direct feedthrough. *Mech Syst Signal Process* 2018; 99: 886–920.

36. Rogers TJ, Worden K and Cross EJ. On the application of gaussian process latent force models for joint input-state-parameter estimation: with a view to Bayesian operational identification. *Mech Syst Signal Process* 2020; 140: 106580.

37. Huang K, Yuen K-V and Wang L. Real-time simultaneous input-state-parameter estimation with modulated colored noise excitation. *Mech Syst Signal Process* 2022; 165: 108378.

38. Teymouri D, Sedehi O, Katafygiotis LS, et al. Input-state-parameter-noise identification and virtual sensing in dynamical systems: a Bayesian expectation-maximization (BEM) perspective. *Mech Syst Signal Process* 2023; 185: 109758.

39. Capalbo CE, Gregoriis DD, Tamarozzi T, et al. Parameter, input and state estimation for linear structural dynamics using parametric model order reduction and augmented Kalman filtering. *Mech Syst Signal Process* 2023; 185: 109799.

40. Impraimakis M and Smyth AW. An unscented Kalman filter method for real time input-parameter-state estimation. *Mech Syst Signal Process* 2022; 162: 108026.

41. Impraimakis M and Smyth AW. A new residual-based Kalman filter for real time input–parameter–state estimation using limited output information. *Mech Syst Signal Process* 2022; 178: 109284.

42. Lin J-W, Betti R, Smyth AW, et al. On-line identification of non-linear hysteretic structural systems using a variable trace approach. *Earthquake Eng Struct Dynamics* 2001; 30(9): 1279–1303.

43. De Angelis M, Luş H, Betti R, et al. Extracting physical parameters of mechanical models from identified state-space representations. *J Appl Mech* 2002; 69(5): 617–625.

44. Sun H, Luş H and Betti R. Identification of structural models using a modified artificial bee colony algorithm. *Comput Struct* 2013; 116: 59–74.

45. Zhi L, QS Li and Fang M. Identification of wind loads and estimation of structural responses of super-tall buildings by an inverse method. *Comput Aided Civ Infrastruct Eng* 2016; 31(12): 966–982.

46. Impraimakis M and Smyth AW. Input-parameter-state estimation of limited information wind-excited systems using a sequential Kalman filter. *Struct Control Health Monit* 2022; 29: 1–16.

47. Brewick PT and Smyth AW. An investigation of the effects of traffic induced local dynamics on global damping estimates using operational modal analysis. *Mech Syst Signal Process* 2013; 41(1–2): 433–453.

48. Papadimitriou C and Argyris C. Bayesian optimal experimental design for parameter estimation and response predictions in complex dynamical systems. *Procedia Eng* 2017; 199: 972–977.

49. Capellari G, Chatzi E and Mariani S. Structural health monitoring sensor network optimization through Bayesian experimental design. *ASCE-ASME J Risk Uncertainty Eng Syst Part A: Civ Eng* 2018; 4(2): 04018016.

50. Ercan T, Sedehi O, Katafygiotis LS, et al. Information theoretic-based optimal sensor placement for virtual sensing using augmented Kalman filtering. *Mech Syst Signal Process* 2023; 188: 110031.

51. Papadimitriou C, Beck JL and Au S-K. Entropy-based optimal sensor location for structural model updating. *J Vibr Control* 2000; 6(5): 781–800.

52. Cumbo R, Mazzanti L, Tamarozzi T, et al. Advanced optimal sensor placement for Kalman-based multiple-input estimation. *Mech Syst Signal Process* 2021; 160: 107830.

53. Flynn EB and Todd MD. A Bayesian approach to optimal sensor placement for structural health monitoring with application to active sensing. *Mech Syst Signal Process* 2010; 24(4): 891–903.

54. Bagirgan B, Mehrjoo A, Moaveni B, et al. Iterative optimal sensor placement for adaptive structural identification using mobile sensors: numerical application to a footbridge. *Mech Syst Signal Process* 2023; 200: 110556.

55. Mehrjoo A, Song M, Moaveni B, et al. Optimal sensor placement for parameter estimation and virtual sensing of strains on an offshore wind turbine considering sensor installation cost. *Mech Syst Signal Process* 2022; 169: 108787.

56. Rainieri C and Fabbrocino G. Automated output-only dynamic identification of civil engineering structures. *Mech Syst Signal Process* 2010; 24(3): 678–695.

57. Bao Y, Tang Z, Li H, et al. Computer vision and deep learning–based data anomaly detection method for structural health monitoring. *Struct Health Monit* 2019; 18(2): 401–421.

58. Atha DJ and Jahanshahi MR. Evaluation of deep learning approaches based on convolutional neural networks

for corrosion detection. *Struct Health Monit* 2018; 17(5): 1110–1128.

59. Yu Y, Wang C, Gu X, et al. A novel deep learning-based method for damage identification of smart building structures. *Struct Health Monit* 2019; 18(1): 143–163.

60. Abdeljaber O, Avci O, Kiranyaz S, et al. Real-time vibration-based structural damage detection using one-dimensional convolutional neural networks. *J Sound Vibr* 2017; 388: 154–170.

61. Abdeljaber O, Avci O, Kiranyaz MS, et al. 1-D CNNs for structural damage detection: verification on a structural health monitoring benchmark data. *Neurocomputing* 2018; 275: 1308–1317.

62. Kolappan Geetha G, Yang H-J and Sim S-H. Fast detection of missing thin propagating cracks during deep-learning-based concrete crack/non-crack classification. *Sensors* 2023; 23(3): 1419.

63. Impraimakis M. A convolutional neural network deep learning method for model class selection. *Earthquake Eng Struct Dynamics* 2024; 53(2): 784–814.

64. Cha Y-J, Choi W and Büyüköztürk O. Deep learning-based crack damage detection using convolutional neural networks. *Comput Aided Civ Infrastruct Eng* 2017; 32(5): 361–378.

65. Zhai G, Narazaki Y, Wang S, et al. Synthetic data augmentation for pixel-wise steel fatigue crack identification using fully convolutional networks. 2022; 29: 237–250.

66. Eren L, Ince T and Kiranyaz S. A generic intelligent bearing fault diagnosis system using compact adaptive 1d cnn classifier. *J Signal Process Syst* 2019; 91: 179–189.

67. Zhang W, Li C, Peng G, et al. A deep convolutional neural network with new training methods for bearing fault diagnosis under noisy environment and different working load. *Mech Syst Signal Process* 2018; 100: 439–453.

68. Avci O, Abdeljaber O, Kiranyaz S, et al. A review of vibration-based damage detection in civil structures: from traditional methods to machine learning and deep learning applications. *Mech Syst Signal Process* 2021; 147: 107077.

69. Guo X, Chen L and Shen C. Hierarchical adaptive deep convolution neural network and its application to bearing fault diagnosis. *Measurement* 2016; 93: 490–502.

70. Janssens O, Slavkovikj V, Vervisch B, et al. Convolutional neural network based fault detection for rotating machinery. *J Sound Vibr* 2016; 377: 331–345.

71. Li S, Liu G, Tang X, et al. An ensemble deep convolutional neural network model with improved ds evidence fusion for bearing fault diagnosis. *Sensors* 2017; 17(8): 1729.

72. Quqa S, Martakis P, Movsessian A, et al. Two-step approach for fatigue crack detection in steel bridges using convolutional neural networks. *J Civ Struct Health Monit* 2022; 12(1): 127–140.

73. Diao Y, Lv J, Wang Q, et al. Structural damage identification based on variational mode decomposition–hilbert transform and CNN. *J Civ Struct Health Monit* 2023; 13: 1–15.

74. Sun S-B, He Y-Y, Zhou S-D, et al. A data-driven response virtual sensor technique with partial vibration measurements using convolutional neural network. *Sensors* 2017; 17(12): 2888.

75. Hubel DH and Wiesel TN. Receptive fields and functional architecture of monkey striate cortex. *J Physiol* 1968; 195(1): 215–243.

76. Fukushima K. Neocognitron: a self-organizing neural network model for a mechanism of pattern recognition unaffected by shift in position. *Biol Cybernet* 1980; 36(4): 193–202.

77. LeCun Y, Boser B, Denker J, et al. Handwritten digit recognition with a back-propagation network. *Adv Neural Inform Process Syst* 1989; 2: 396–404.

78. Kolen JF and Kremer SC. *A field guide to dynamical recurrent networks*. New York, NY: John Wiley & Sons, 2001.

79. Houdt GV, Mosquera C and Nápoles G. A review on the long short-term memory model. *Artif Intell Rev* 2020; 53: 5929–5955.

80. Hochreiter S and Schmidhuber J. Long short-term memory. *Neural Comput* 1997; 9(8): 1735–1780.

81. Zhou JM, Dong L, Guan W, et al. Impact load identification of nonlinear structures using deep recurrent neural network. *Mech Syst Signal Process* 2019; 133: 106292.

82. Kamariotis A, Tatsis K, Chatzi E, et al. A metric for assessing and optimizing data-driven prognostic algorithms for predictive maintenance. *Reliab Eng Syst Saf* 2024; 242: 109723.

83. Xu Y, Lu X, Cetiner B, et al. Real-time regional seismic damage assessment framework based on long short-term memory neural network. *Comput Aided Civ Infrastruct Eng* 2021; 36(4): 504–521.

84. Cho K, Merriënboer BV, Gulcehre C, et al. Learning phrase representations using RNN encoder-decoder for statistical machine translation. *arXiv preprint arXiv:1406.1078*, 2014.

85. Dey R and Salem FM. Gate-variants of gated recurrent unit (GRU) neural networks. In: *2017 IEEE 60th international midwest symposium on circuits and systems (MWSCAS)*, 2017, pp. 1597–1600. IEEE.

86. Fu D, Wang L, Lv G, et al. Advances in dynamic load identification based on data-driven techniques. *Eng Appl Artif Intell* 2023; 126: 106871.

87. Zhang C, Zhang W, Ying G, et al. A deep learning method for heavy vehicle load identification using structural dynamic response. *Comput Struct* 2024; 297: 107341.

88. Zheng Z, Yi C, Lin J, et al. A novel deep learning architecture and its application in dynamic load monitoring of the vehicle system. *Measurement* 2024; 229: 114336.

89. Moradi S, Duran B, Azam SE, et al. Novel physics-informed artificial neural network architectures for system and input identification of structural dynamics pdes. *Buildings* 2023; 13(3): 650.

90. Guo C, Jiang L, Yang F, et al. An intelligent impact load identification and localization method based on autonomic feature extraction and anomaly detection. *Eng Struct* 2023; 291: 116378.

91. Liu Y, Wang L and Gu K. A support vector regression (SVR)-based method for dynamic load identification using heterogeneous responses under interval uncertainties. *Appl Soft Comput* 2021; 110: 107599.

92. Zhang X, He W, Cui Q, et al. Wavloadnet: dynamic load identification for aeronautical structures based on convolution neural network and wavelet transform. *Appl Sci* 2024; 14(5): 1928.

93. Liu Y, Wang L, Gu K, et al. Artificial neural network (ANN)-Bayesian probability framework (BPF) based method of dynamic force reconstruction under multisource uncertainties. *Knowl Based Syst* 2022; 237: 107796.

94. He W, Zhang X, Feng Z, et al. Random dynamic load identification with noise for aircraft via attention based 1D-CNN. *Aerospace* 2022; 10(1): 16.

95. Khosrowpour E and Hematiyan MR. Distributed load identification for hyperelastic plates using gradient-based and machine learning methods. *Acta Mech* 2024; 235: 3271–3291.

96. Yang H, Jiang J, Chen G, et al. Dynamic load identification based on deep convolution neural network. *Mech Syst Signal Process* 2023; 185: 109757.

97. Yang H, Jiang J, Chen G, et al. A recurrent neural network-based method for dynamic load identification of beam structures. *Materials* 2021; 14(24): 7846.

98. Baek SM, Park JC and Jung HJ. Impact load identification method based on artificial neural network for submerged floating tunnel under collision. *Ocean Eng* 2023; 286: 115641.

99. Wang L, Huang Y, Xie Y, et al. A new regularization method for dynamic load identification. *Sci Progress* 2020; 103(3): 0036850420931283.

100. Holmes G, Sartor P, Reed S, et al. Prediction of landing gear loads using machine learning techniques. *Struct Health Monit* 2016; 15(5): 568–582.

101. Zhao Y, Noori M, Altabey WA, et al. Deep learning-based damage, load and support identification for a composite pipeline by extracting modal macro strains from dynamic excitations. *Appl Sci* 2018; 8(12): 2564.

102. Feng T, Duan A, Guo L, et al. Deep learning based load and position identification of complex structure. In: *2021 IEEE 16th conference on industrial electronics and applications (ICIEA)*, 2021, pp. 1358–1363. IEEE.

103. Guo C, Jiang L, Yang F, et al. Impact load identification and localization method on thin-walled cylinders using machine learning. *Smart Mater Struct* 2023; 32(6): 065018.

104. Gai T, Yu D, Zeng S, et al. An optimization neural network model for bridge cable force identification. *Eng Struct* 2023; 286: 116056.

105. Ren-Mu H and Germond AJ. Comparison of dynamic load modeling using neural network and traditional method. In: [1993] *Proceedings of the second international forum on applications of neural networks to power systems*, Yokohama, Japan, 1992, pp. 253–258. IEEE.

106. Rosafalco L, Manzoni A, Mariani S, et al. An autoencoder-based deep learning approach for load identification in structural dynamics. *Sensors* 2021; 21(12): 4207.

107. Yuen K-V, Liang P-F and Kuok S-C. Online estimation of noise parameters for Kalman filter. *Struct Eng Mech* 2013; 47(3): 361–381.

108. Impraimakis M. A kullback–leibler divergence method for input–state–state identification. *J Sound Vib* 2024; 569: 117965.

109. Kingma DP and Ba J. Adam: a method for stochastic optimization. *arXiv preprint arXiv:1412.6980*, 2014.

110. Rao C, Sun H and Yang Liu. Physics-informed deep learning for computational elastodynamics without labeled data. *J Eng Mech* 2021; 147(8): 04021043.

111. Wang Z-W, Lu X-F, Zhang W-M, et al. Deep learning-based reconstruction of missing long-term girder-end displacement data for suspension bridge health monitoring. *Comput Struct* 2023; 284: 107070.

112. Haddadi H, Shakal A, Stephens C, et al. Center for engineering strong-motion data (CESMD). In: *Proceedings of the 14th World conference on earthquake engineering*, Beijing, October, pp. 12–17, 2008.

113. Maes K, Chatzis MN and Lombaert G. Observability of nonlinear systems with unmeasured inputs. *Mech Syst Signal Process* 2019; 130: 378–394.

114. Eftekhar Azam S, Chatzi E, Papadimitriou C, et al. Experimental validation of the Kalman-type filters for online and real-time state and input estimation. *J Vibr Control* 2017; 23(15): 2494–2519.

115. Dyke SJ, Bernal D, Beck J, et al. Experimental phase ii of the structural health monitoring benchmark problem. In: *Proceedings of the 16th ASCE engineering mechanics conference*, 2003.

116. Huang M, Cheng X and YongZhi Lei. Structural damage identification based on substructure method and improved whale optimization algorithm. *J Civ Struct Health Monit* 2021, 11: 351–380.

117. Giraldo DF, Song W, Dyke SJ, et al. Modal identification through ambient vibration: comparative study. *J Eng Mech* 2009; 135(8): 759–770.

118. Jian-ye Ching and Beck JL. Bayesian analysis of the phase II IASC–ASCE structural health monitoring experimental benchmark data. *J Eng Mech* 2004; 130(10): 1233–1244.

119. Raissi M, Perdikaris P and Karniadakis GE. Physics-informed neural networks: a deep learning framework for solving forward and inverse problems involving nonlinear partial differential equations. *J Comput Phys* 2019; 378: 686–707.

120. Zhu Y, Zabaras N, Koutsourelakis P-S, et al. Physics-constrained deep learning for high-dimensional surrogate modeling and uncertainty quantification without labeled data. *J Comput Phys* 2019; 394: 56–81.

121. Olivier A, Mohammadi S, Smyth AW, et al. Bayesian neural networks with physics-aware regularization for probabilistic travel time modeling. *Comput Aided Civ Infrastruct Eng* 2023; 38: 2614–2631.

122. Lai Z, Liu W, Jian X, et al. Neural modal ordinary differential equations: integrating physics-based modeling with neural ordinary differential equations for modeling high-dimensional monitored structures. *Data-Centric Eng* 2022; 3: e34.

123. Lai Z, Mylonas C, Nagarajaiah S, et al. Structural identification with physics-informed neural ordinary differential equations. *J Sound Vibr* 2021; 508: 116196.

124. Zhang R, Liu Y and Sun H. Physics-guided convolutional neural network (PhyCNN) for data-driven seismic response modeling. *Eng Struct* 2020; 215: 110704.

125. Chen Z, Liu Y and Sun H. Physics-informed learning of governing equations from scarce data. *Nat Commun* 2021; 12(1): 6136.

126. Eshkevari SS, Takáč M, Pakzad SN, et al. Dynnet: physics-based neural architecture design for nonlinear structural response modeling and prediction. *Eng Struct* 2021; 229: 111582.

127. Chatzis MN, Chatzi EN and Smyth AW. On the observability and identifiability of nonlinear structural and mechanical systems. *Struct Control Health Monit* 2015; 22(3): 574–593.

128. Shi X, Williams MS and Chatzis MN. A robust algorithm to test the observability of large linear systems with unknown parameters. *Mech Syst Signal Process* 2021; 157: 107633.

129. Eshkevari SS, Cronin L, Eshkevari SS, et al. Input estimation of nonlinear systems using probabilistic neural network. *Mech Syst Signal Process* 2022; 166: 108368.

130. Sarego G, Zaccariotto M and Galvanetto U. Artificial neural networks for impact force reconstruction on composite plates and relevant uncertainty propagation. *IEEE Aerospace Electron Syst Mag* 2018; 33(8): 38–47.

131. Wang L, Liu Y, Gu K, et al. A radial basis function artificial neural network (RBF ANN) based method for uncertain distributed force reconstruction considering signal noises and material dispersion. *Comput Methods Appl Mech Eng* 2020; 364: 112954.